\documentclass[letterpaper]{article} %
\usepackage{aaai2027}  %
\nocopyright
\usepackage{amsmath}
\usepackage{amssymb}
\usepackage{dsfont}
\usepackage{multirow}
\usepackage[hyphens]{url}  %
\usepackage{graphicx} %
\usepackage{natbib}  %
\usepackage{caption} %
\usepackage{booktabs}   %
\usepackage{siunitx}    %
\newcommand{\indicator}{\mathbf{1}}
\newenvironment{promptbox}[1]{%
  \begin{quote}\noindent\textbf{#1}\par
}{%
  \end{quote}
}
\usepackage{algorithm}
\usepackage{algorithmic}
\usepackage{enumitem}

\usepackage{newfloat}
\usepackage{listings}
\DeclareCaptionStyle{ruled}{labelfont=normalfont,labelsep=colon,strut=off} %
\floatstyle{ruled}
\newfloat{listing}{tb}{lst}{}
\floatname{listing}{Listing}
\newcommand{\methodname}{\textsc{ToolLIFT}}

\title{\methodname{}: Lifting Tool-Specific Trajectories into Function-Level Graphs for Generalizable Tool Planning}
\author{
    Xiuhui You\textsuperscript\rm{,}
    Jiayi Luo\textsuperscript\rm{,}
    Zichao Shen\textsuperscript\rm{,}
    Qingyun Sun\textsuperscript\rm{,}
    Ziwei Zhang\textsuperscript{*}
}
\affiliations{
School of Computer Science and Engineering\\
Beihang University\\
\{xiuhuiyou,zwzhang\}@buaa.edu.cn
}
\begin{document}

\maketitle

\begin{abstract}
Historical tool-use trajectories provide valuable experience for large language model (LLM) agents to plan and coordinate tool usage. 
Existing approaches directly construct tool-level graphs from these
trajectories, but the resulting graphs remain tied to specific tools and are hard to generalize across tool sets.
To tackle this challenge, we find that despite differences in the tools involved, analogous tasks often share a common function-level workflow structure, which serves as a potentially more transferable abstraction for tool planning.
Based on this insight, we propose \textbf{\methodname{}}, a framework that lifts tool-specific trajectories into a function-level workflow graph (FWG) for generalizable tool planning.  
Specifically, we first propose a trajectory-lifting mechanism that encodes workflow structures in the FWG and shares collaboration experience across tools.
Then, building on the global structure of the FWG, we introduce decoupled workflow planning and tool selection to align individual tool choices with the overall workflow.
Lastly, to ensure reliable tool dataflow, we adopt Reinforcement Learning (RL) and propose source-gated and skill-specific rewards to maintain source-traceable information flow across tool calls.
Experiments on two in-distribution (ID) and three
out-of-distribution (OOD) benchmarks show that \methodname{} consistently outperforms state-of-the-art baselines, 
demonstrating strong generalization to unseen tool sets.
\end{abstract}

\section{Introduction}
Large language model (LLM) agents can reuse historical experience to tackle complex real-world tasks rather than reason from scratch~\cite{wang2024survey,toolnet,toolexpnet}.
Tool-use trajectories provide a natural record of this experience, capturing how multiple tools collaborate through invocation sequences and data dependencies. Therefore, aggregating collaboration experience from individual trajectories into a reusable structure has become an important research direction for LLM agents.

\begin{figure}[t]
    \centering
    \includegraphics[width=\linewidth]{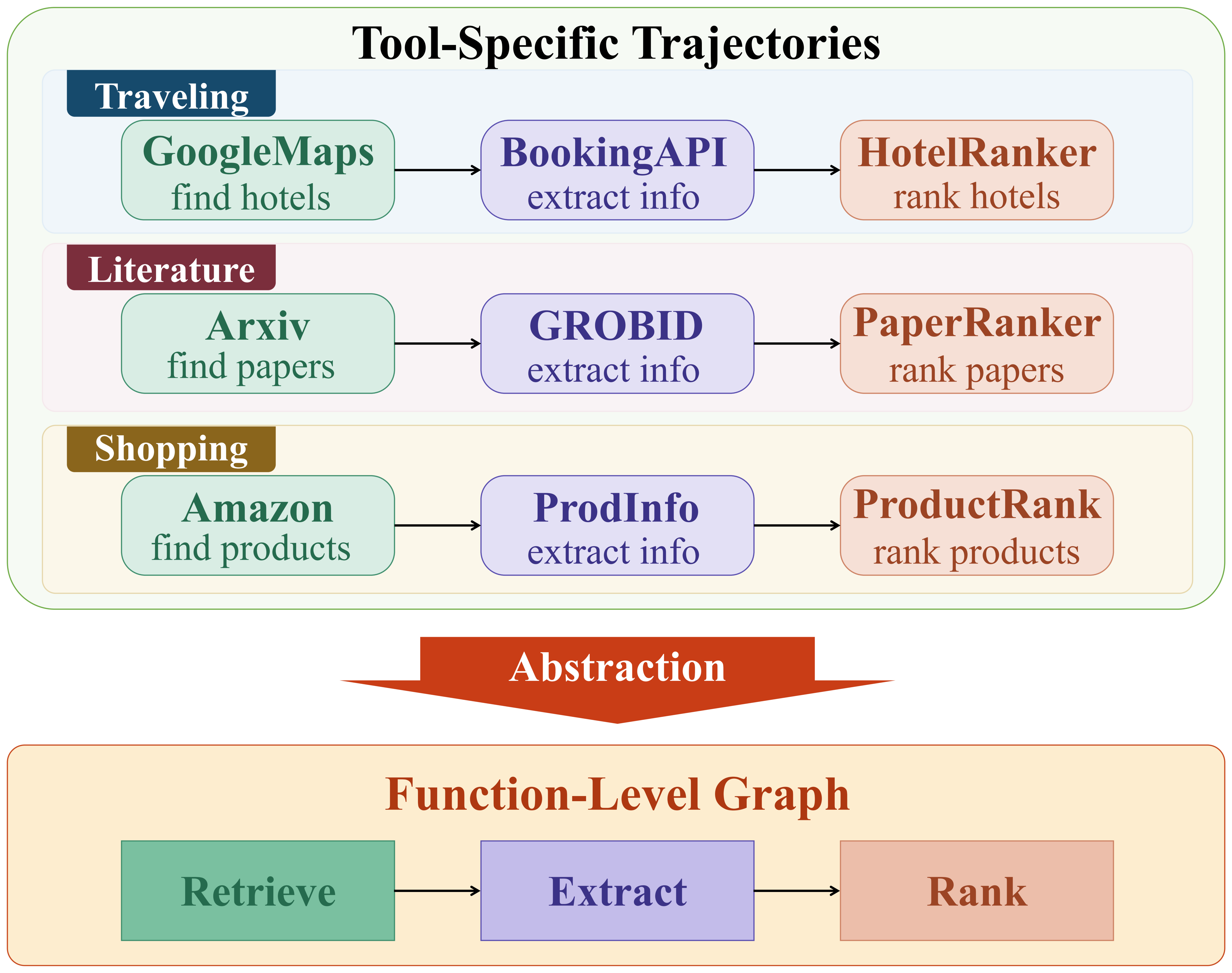}
    \caption{Tool-use trajectories instantiated with different tool sets can share a function-level workflow structure: although the concrete tools differ, their functional roles and relations follow the same pattern.}
    \label{fig:insight}
\end{figure}

Recent work constructs tool-level graphs~\cite{toolnet,gtool,naviagent,autotool,toolexpnet}, where nodes denote tools and edges encode relations derived from historical usage. However, these graphs bind collaboration experience only to tools observed in historical trajectories, limiting transfer to unseen tool sets. Besides, existing methods typically construct plans by searching step by step over local tool transitions in the graph~\cite{autotool,wang2023plan,naviagent}, without
establishing a globally consistent tool plan.

These limitations raise three fundamental questions in graph-based tool planning:
(1) \emph{How can collaboration experience transfer to tools in unseen tool sets?}
Tool-level graphs generalize poorly as rarely used tools are sparsely connected, while unseen tools from other tool sets may be entirely isolated from the graph.
(2) \emph{How can an agent maintain a global view when choosing individual tools?}
Stepwise planning commits to concrete tools without a complete view of the workflow structure, potentially leading to myopic choices that cause the overall plan to fail.
During planning, an agent must not only determine the workflow structure but also fill tool arguments, which requires tracking argument-level dataflow across tool calls.
(3) \emph{How can an agent maintain reliable dataflow across tool calls?}
As intermediate tool outputs accumulate in the context, identifying the correct source of each tool argument becomes increasingly difficult. This could increase the risk of hallucinated values or incorrect dependencies~\cite{patil2024gorilla,song2023restgpt,toolrl,toolzero}.

To address the aforementioned challenges, our key insight, as illustrated in Figure~\ref{fig:insight}, is that tool-use trajectories instantiated with different tool sets for analogous tasks often share the same function-level workflow structure, which serves as a potentially more transferable abstraction across concrete tool sets. Although the concrete tools at corresponding positions may differ, they serve the same functional roles.

Based on this insight, we propose \textbf{\methodname{}}, a framework that lifts tool-specific trajectories into function-level graphs for generalizable tool planning. Specifically, we first propose Function-Level Workflow Graph Construction via Trajectory Lifting, aiming to make historical tool-use experience reusable beyond concrete tool identities. Our method constructs a Function-Level Workflow Graph (FWG) by lifting concrete tools to functions based on their functional features and aggregating tool transitions into function-level transitions. The FWG allows each tool to benefit from collaboration structures learned from all tools sharing the same function. 
Second, to keep individual tool choices consistent with the global workflow, we introduce Decoupled Workflow Planning and Tool Selection, which leverages the FWG to plan the complete workflow and then instantiates each function with a concrete tool under the corresponding function constraints. 
Finally, to maintain reliable dataflow across tool calls, we introduce RL-Based Dataflow Learning for Source-Traceable Tool Calls, which explicitly determines whether each argument is a direct value from the context or a reference to a preceding tool output. To learn these dependencies, we adopt reinforcement learning (RL) via Group Relative Policy Optimization (GRPO)~\cite{grpo}, using source-gated and skill-specific argument rewards that first validate the source type and then provide separate signals for argument filling and source tracing.

Experiments on two ID and three OOD benchmarks show that \methodname{} consistently outperforms state-of-the-art baselines, with stronger gains in OOD settings. These results demonstrate improvements in tool planning and generalization to unseen tool sets.

We summarize the contributions as follows:
\begin{itemize}[nosep]
    \item We propose \methodname{}, a framework that lifts tool-specific trajectories into a Function-Level Workflow Graph, making collaboration experience transferable across tool sets.
    \item We introduce decoupled workflow planning and tool selection to align individual choices with the global workflow, together with RL-Based Dataflow Learning to maintain source-traceable information flow across tool calls.
    \item Experiments using different LLM backbones across two ID and three OOD benchmarks demonstrate consistent performance and strong generalization to unseen tool sets.
\end{itemize}

\section{Related Work}

\subsection{Experience Reuse for Agent Tool Use}

LLM agents can reuse experience through retrieved external
knowledge~\cite{lewis2020retrieval}, agent memory and reusable
guidelines~\cite{park2023generative,shinn2023reflexion,
fu2024autoguide,wang2025agent}, or skill
libraries~\cite{zhao2024expel,wang2024voyager,wang2026skillx}.
For tool calling, tool-use trajectories offer direct tool invocation experience: RestGPT uses in-context examples to guide API planning and selection~\cite{song2023restgpt}, while ToolACE synthesizes function-calling trajectories for supervised fine-tuning~\cite{liu2025toolace}.
However, these methods preserve trajectories in tool-specific form, leaving
their collaboration patterns tied to concrete tools and difficult to share
across tool sets.

\subsection{Graph-Based and Hierarchical Tool Planning}

Modeling inter-tool collaboration patterns as a tool graph has emerged as an
effective way to guide tool planning.
Schema-based methods derive graph edges from input-output compatibility; ControlLLM, for example, searches the resulting graph to identify composable tool chains~\cite{controlllm}.
Experience-driven methods such as ToolExpNet instead construct graphs using
transition dependencies extracted from historical trajectories~\cite{toolexpnet}.
However, these graphs remain tied to concrete tool identities, limiting reuse
across tool sets.
Tool-Planner alleviates this limitation by grouping similar APIs into toolkits to support runtime tool substitution, but does not separate functional and domain features or distill reusable transition patterns from historical trajectories~\cite{toolplanner}.
Hierarchical and workflow-based methods further separate high-level planning
from concrete execution. 
HuggingGPT plans abstract tasks and resource dependencies before model
selection, ReWOO separates planning from execution through evidence variables,
and NaviAgent performs bilevel planning over a tool navigation
graph~\cite{shen2023hugginggpt,xu2023rewoo,naviagent}.
However, their high-level plans are primarily formed through semantic task
decomposition, with tool structures mainly supporting subsequent selection or
navigation.
In contrast, \methodname{} grounds high-level workflow planning in reusable
collaboration structures distilled from prior tool use before realizing the
workflow with concrete tools.

\subsection{Reinforcement Learning for Tool Planning}

Recent work applies reinforcement learning to tool-integrated reasoning and
multi-step tool use~\cite{feng2025retool,jin2025searchr1,yu2025steptool}.
For multi-tool settings, Tool-Star trains models to coordinate multiple tools
during stepwise reasoning~\cite{dong2025toolstar}, while ToolRL decomposes
correctness into fine-grained rewards for tool names, parameter names, and
parameter values~\cite{toolrl}.
However, these rewards do not explicitly supervise where each argument value
comes from. Consequently, argument-source correctness is not directly
optimized.

\begin{figure*}[!t]
    \centering
    \includegraphics[width=\textwidth]{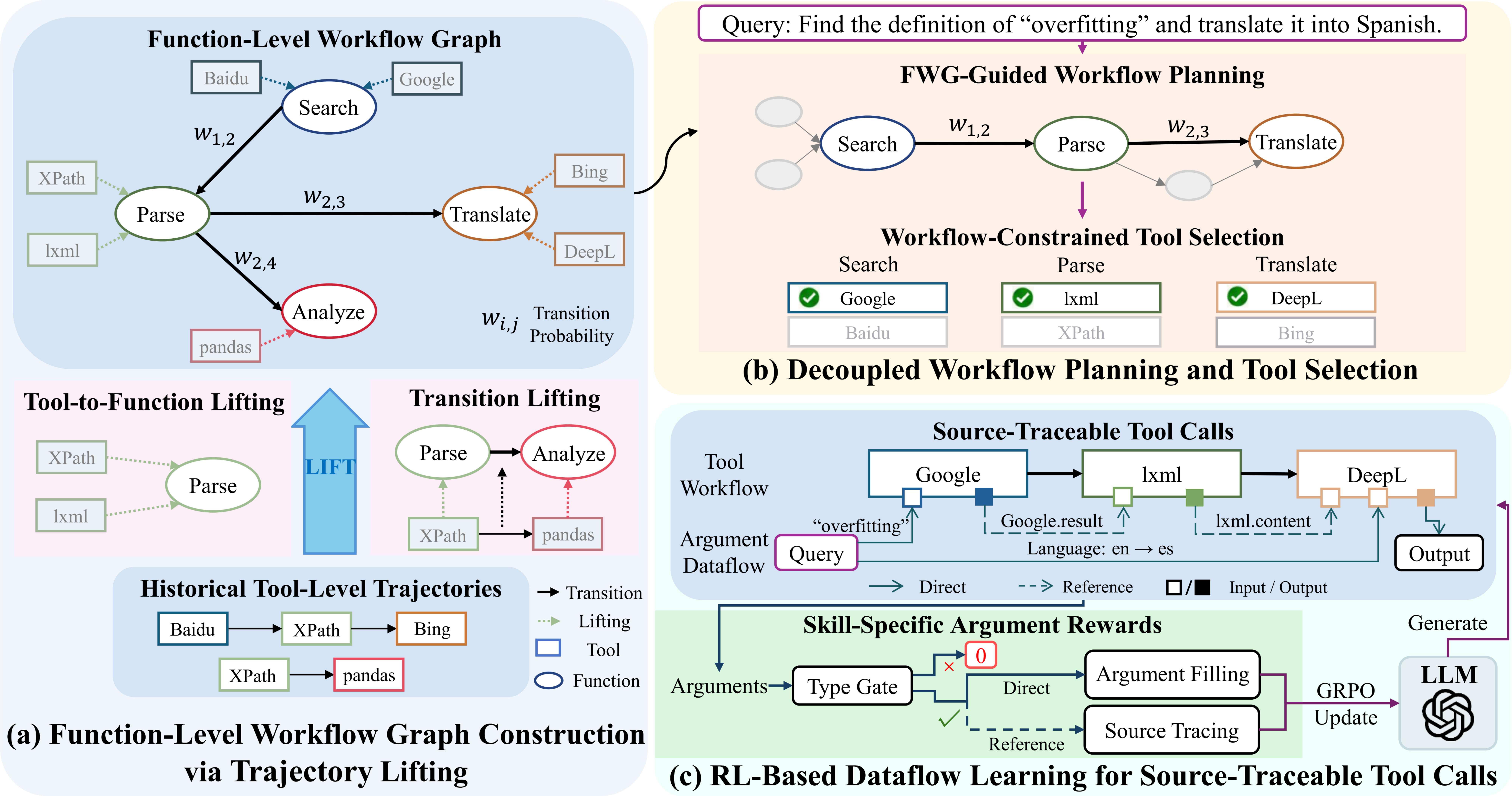}
    \caption{An overview of \methodname{}.
    (a) We lift tool-specific trajectories into a Function-Level Workflow Graph (FWG) that captures function-level collaboration structures shared across tool sets.
    (b) We plan a function-level workflow over the global structure of the FWG and instantiate it with concrete tools under constraints, keeping individual tool choices aligned with the overall workflow.
    (c) We explicitly assign argument sources for source-traceable
    dataflow and jointly optimize argument filling and source tracing with
    source-gated and skill-specific rewards.}
    \label{fig:method}
\end{figure*}
\section{Problem Definition}
\label{sec:prelim}

We consider an LLM agent equipped with a tool set $\mathcal{T}$. Each tool $t\in\mathcal{T}$ has a textual schema $d_t$ that describes its functionality and input and output arguments.

\paragraph{Historical trajectories.}
We denote the corpus of historical tool-use trajectories by $\mathcal{D}=\{(q^{(m)},\mathcal{W}_t^{(m)})\}_{m=1}^{M}$, where $\mathcal{W}_t^{(m)}=(t_1^{(m)},\ldots,t_{K_m}^{(m)})$ is the $m$-th trajectory of length $K_m$ that solves the query $q^{(m)}$.

\paragraph{Tool-planning task.}
Given a query $q$ and candidate tools $\mathcal{T}_q\subseteq\mathcal{T}$, tool planning produces an ordered tool-call plan $\mathcal{P}=(p_1,p_2,\ldots,p_K)$ where each call $p_k=(t_k,\mathcal{A}_k)$ specifies a tool $t_k\in\mathcal{T}_q$ and its input argument assignments $\mathcal{A}_k$. An argument value can be supplied by the query context or refer to the output of an earlier call $p_j$ with $j<k$. We assume that each argument has a single source: either the query context or one preceding tool call. 

\section{Method}
\label{sec:method}
 
Figure~\ref{fig:method} illustrates \methodname{}. Our method has three components. First, we build an FWG that captures function-level workflow structures and enables collaboration experience to be
shared across tool sets (Section~\ref{sec:fwg}). Next, we use the global FWG
structure to establish a functionally complete plan and instantiate its functional roles with concrete tools (Section~\ref{sec:planning}). Finally, we model argument-level dependencies for source-traceable information flow across tool calls (Section~\ref{sec:dataflow}). We optimize the framework in two stages using GRPO with rule-based rewards (Section~\ref{sec:grpo}).

\subsection{Function-Level Workflow Graph Construction via Trajectory Lifting}
\label{sec:fwg}

Given a training tool set $\mathcal{T}_{\mathrm{train}}$ and its historical
trajectory corpus $\mathcal{D}_{\mathrm{train}}$, we construct an FWG by
lifting tools and tool transitions into functions and function-level
transitions, respectively, as illustrated in Figure~\ref{fig:method}(a).

\paragraph{Tool-to-Function Lifting.}
To group concrete tools by their abstract functions, we cluster them using
functional features. However, directly embedding a complete tool schema may
overemphasize domain-specific features, thereby obscuring the underlying
function. To separate functional features from domain-specific information,
we prompt an LLM to decompose each training tool's schema $d_t$ into a
functional description $d_t^{\mathrm{func}}$ and a domain description
$d_t^{\mathrm{dom}}$. This separation allows tools from different domains to be compared by function rather than domain-specific content, enabling tools that serve the same role to be grouped together. We encode only
$d_t^{\mathrm{func}}$ with BGE-M3~\cite{bgem3} and fit UMAP~\cite{umap} to the
embeddings of the training tools
${\mathbf{e}*t}*{t \in \mathcal{T}_{\mathrm{train}}}$, obtaining compact
representations $\tilde{\mathbf{e}}*t$. We then apply
$K$-means~\cite{macqueen1967some} to
${\tilde{\mathbf{e}}*t}*{t \in \mathcal{T}*{\mathrm{train}}}$, obtaining $L$
functional clusters $\mathcal{C}={c_1,\ldots,c_L}$. We select $L$ by
maximizing the silhouette coefficient~\cite{rousseeuw1987silhouettes}, which
measures both within-cluster cohesion and separation between clusters;
sensitivity to $L$ is analyzed in Section~\ref{sec:rq3}. The resulting mapping
$\phi: \mathcal{T} \rightarrow \mathcal{C}$ assigns each tool to an abstract
function.

\paragraph{Trajectory Lifting.}
Using $\phi$, each tool-level trajectory
$\mathcal{W}_t^{(m)}=(t_1^{(m)},\ldots,t_{K_m}^{(m)})$ is lifted into the
function-level workflow
$\mathcal{W}_c^{(m)}=(c_1^{(m)},\ldots,c_{K_m}^{(m)})$, where
$c_k^{(m)}=\phi(t_k^{(m)})$. We count adjacent function pairs across all $M$ workflows:
\begin{equation} 
n(c \to c') = \sum_{m=1}^{M} \sum_{k=1}^{K_m - 1} 
\mathds{1}\!\left[c_k^{(m)} = c,\; c_{k+1}^{(m)} = c'\right]. 
\end{equation}
We row-normalize these counts into transition probabilities:
\begin{equation}
w(c, c') = \frac{n(c \to c')}{\sum_{c'' \in \mathcal{C}} 
n(c \to c'')}.
\label{eq:transition}
\end{equation}
We retain self-transitions ($c=c'$) because a functional role may recur within
a workflow. These transitions define the directed weighted graph
\begin{equation}
\mathcal{G}_{\text{fwg}} = (\mathcal{C}, \mathcal{E}, w), 
\quad \mathcal{E} = \{(c, c') : n(c \to c') > 0\},
\end{equation}
where each edge $(c,c')\in\mathcal{E}$ carries weight $w(c,c')\in[0,1]$. By aggregating transitions at the function level, the FWG captures
collaboration structures shared across concrete tools.

\paragraph{Cold-Start Transition Inheritance.}
For an unseen tool $t_{\mathrm{new}}$, we apply the same lifting procedure and
assign it to the functional cluster with the nearest centroid:
\begin{equation}
\phi(t_{\text{new}}) = \arg\min_{c \in \mathcal{C}} 
\|\tilde{\mathbf{e}}_{t_{\text{new}}} - \boldsymbol{\mu}_c\|_2,
\end{equation}
where $\boldsymbol{\mu}_c$ is the centroid of cluster $c$. The new tool then
inherits the FWG transitions of its function.

\subsection{Decoupled Workflow Planning and Tool Selection}
\label{sec:planning}

To align individual tool choices with the global workflow, we decouple
workflow planning from tool selection, as illustrated in Figure~\ref{fig:method}(b). The planner
uses FWG transitions as soft guidance to generate a functionally complete workflow,
which the tool-call generator instantiates with concrete tools under the
corresponding constraints.

\paragraph{FWG-Guided Workflow Planning.}
Given a query $q$ and candidate tools
$\mathcal{T}_q\subseteq\mathcal{T}$, we map them to functional roles
$\mathcal{C}_q=\{\phi(t):t\in\mathcal{T}_q\}$ and extract the induced FWG
subgraph $\mathcal{G}_q=(\mathcal{C}_q,\mathcal{E}_q,w)$, where
$\mathcal{E}_q=\mathcal{E}\cap(\mathcal{C}_q\times\mathcal{C}_q)$. We serialize
$\mathcal{G}_q$ as text by listing each function's outgoing neighbors in
descending order of $w(c,c')$. This provides soft evidence for plausible
transitions without determining the next step. Rather than traversing the graph
step by step, the planner conditions on $q$ and $\mathcal{G}_q$ to generate a
complete function-level workflow in one call,
$\mathcal{W}_c=(c_{\sigma(1)},c_{\sigma(2)},\ldots,c_{\sigma(K)})$, where
$\sigma(k)$ indexes the function selected at step $k$.

\paragraph{Function-Coverage Reward.}
A task may admit multiple valid workflow linearizations, particularly when
some calls are independent, so we do not require predictions to follow the annotated order. 
Instead, we compare the predicted and target workflows
$\hat{\mathcal{W}}_c$ and $\mathcal{W}_c^{*}$ as function multisets. For any
workflow $\mathcal{W}$, let $n_{\mathcal{W}}(c)$ denote the number of
occurrences of function $c$. The multiset overlap between them is
\begin{equation}
I(\hat{\mathcal{W}}_c,\mathcal{W}_c^{*})
= \sum \nolimits_{c\in\mathcal{C}} \min\!\left(n_{\hat{\mathcal{W}}_c}(c),
n_{\mathcal{W}_c^{*}}(c)\right),
\end{equation}
and the corresponding multiset Jaccard similarity is
\begin{equation}
\mathrm{MJ}(\hat{\mathcal{W}}_c,\mathcal{W}_c^{*})
=
\frac{I(\hat{\mathcal{W}}_c,\mathcal{W}_c^{*})}
{|\hat{\mathcal{W}}_c| + |\mathcal{W}_c^{*}| -
I(\hat{\mathcal{W}}_c,\mathcal{W}_c^{*})}.
\end{equation}
We combine similarity with the number of matched function instances
and scale the reward to $[-\rho, \rho]$, where $\rho > 0$:
\begin{equation}
R_{\mathrm{func}}^{(1)}
= 2\rho \cdot
\frac{\mathrm{MJ}(\hat{\mathcal{W}}_c,\mathcal{W}_c^{*}) +
I(\hat{\mathcal{W}}_c,\mathcal{W}_c^{*})}
{1 + |\mathcal{W}_c^{*}|} - \rho. 
\end{equation}
This reward provides continuous and informative feedback for partial recovery of the required functional composition while tolerating alternative linearizations. 

\paragraph{Workflow-Constrained Tool Selection.}
Given $\mathcal{W}_c$, the tool-call generator receives $q$ and tool
schemas organized by functional role. For each $c_{\sigma(k)}$, it selects a
tool from the corresponding cluster, producing
$\hat{\mathcal{W}}_t=(\hat{t}_1,\hat{t}_2,\ldots,\hat{t}_K)$ with
$\phi(\hat{t}_k)=c_{\sigma(k)}$. This constraint narrows the selection space and
aligns each tool with its functional role.

\paragraph{Tool Matching Reward.}
We adapt the tool matching reward from ToolRL~\cite{toolrl}. Let
$\hat{\mathcal{P}}$ and $\mathcal{P}^{*}$ be the predicted and target plans.
For finite sets $\mathcal{X}$ and $\mathcal{Y}$, their Jaccard similarity is
\begin{equation}
\mathrm{J}(\mathcal{X},\mathcal{Y})
=
\frac{|\mathcal{X}\cap\mathcal{Y}|}
     {|\mathcal{X}\cup\mathcal{Y}|}.
\end{equation}
We form a matched-call relation $\mathcal{R}$ by pairing predicted and target
calls that invoke the same tool, and compare their argument-name sets
$\operatorname{args}(p)$. The reward combines the tool-set and argument-set
similarities:
\begin{small}
\begin{equation}
r_{\mathrm{match}}
=
\mathrm{J}\!\left(
\operatorname{tools}(\hat{\mathcal{P}}),
\operatorname{tools}(\mathcal{P}^{*})
\right) +
\sum_{(\hat{p},p^{*})\in\mathcal{R}}
\mathrm{J}\!\left(
\operatorname{args}(\hat{p}),
\operatorname{args}(p^{*})
\right).
\end{equation}
\end{small}

\subsection{RL-Based Dataflow Learning for Source-Traceable Tool Calls}
\label{sec:dataflow}

Workflow planning determines the required functional roles and tool-call
linearization, but reliable execution also requires identifying each
argument's source. Inferring argument values from an LLM's parametric knowledge and a growing context may lead the model to confuse their sources or hallucinate values. We therefore represent each argument as either a direct context value or a reference to a preceding tool output, explicitly identifying its
information source at the call level. We learn this explicit dataflow with source-gated and skill-specific argument rewards, as shown in Figure~\ref{fig:method}(c).

Let $\mathcal{P}=(p_1,p_2,\ldots,p_K)$ be the ordered tool-call sequence, where
$p_i=(t_i,\mathcal{A}_i)$ contains the selected tool $t_i$ and its input
arguments $\mathcal{A}_i$. Each argument $a\in\mathcal{A}_i$ takes one of two
dependency forms:
\begin{itemize}
    \item a \textbf{direct argument} $a\leftarrow x$, where $x$ is supplied by
    the query or static context;
    \item a \textbf{reference argument}
    $a\leftarrow\operatorname{out}(p_j)$, where $j<i$ and the value is provided
    by a preceding tool call $p_j$.
\end{itemize}
The constraint $j<i$ captures dataflow precedence without imposing a unique
order on independent calls. During generation, direct arguments are filled
immediately, whereas reference arguments point to preceding tool outputs.
Together, they provide a dataflow blueprint for execution. The FWG provides soft transition guidance, whereas references to preceding tool outputs explicitly encode cross-call precedence.

\paragraph{Skill-Specific Argument Rewards.}
For predicted and target argument values $\hat{v}$ and $v^{*}$, the reward
first applies a source-type gate: a mismatch receives zero reward, while
type-matched values are evaluated with type-specific scores:
\begin{equation}
r_{\mathrm{val}}(\hat{v},v^{*})
=
\begin{cases}
0,
& \tau(\hat{v})\ne\tau(v^{*}),\\
\mathds{1}\!\left[\hat{v}=v^{*}\right],
& \tau(v^{*})=\mathrm{reference},\\
\mathrm{ROUGE\text{-}L}_{\mathrm{F1}}(\hat{v},v^{*}),
& \tau(v^{*})=\mathrm{direct},
\end{cases}
\end{equation}
where $\tau(v)\in\{\mathrm{direct},\mathrm{reference}\}$ denotes the source
type. For direct values, $\mathrm{ROUGE\text{-}L}_{\mathrm{F1}}$~\cite{rouge} with
whitespace tokenization is computed as
\begin{equation}
\mathrm{ROUGE\text{-}L}_{\mathrm{F1}}(\hat{v},v^{*})
=
\frac{
2\left|
\operatorname{LCS}\!\left(
\operatorname{tok}(\hat{v}),
\operatorname{tok}(v^{*})
\right)
\right|
}{
|\operatorname{tok}(\hat{v})|
+
|\operatorname{tok}(v^{*})|
},
\end{equation}
where $\operatorname{tok}(\cdot)$ tokenizes on whitespace and
$\operatorname{LCS}(\cdot,\cdot)$ returns the longest common subsequence.
For each shared argument $a$, let $\hat{v}_a$ and $v_a^{*}$ be its predicted
and target values. The argument-value reward aggregates over matched calls:
\begin{equation}
r_{\mathrm{value}}
=
\sum_{(\hat{p},p^{*})\in\mathcal{R}}
\sum_{a\in
\operatorname{args}(\hat{p})\cap\operatorname{args}(p^{*})}
r_{\mathrm{val}}(\hat{v}_a,v_a^{*}).
\end{equation}
Direct values may admit different valid textual realizations, so ROUGE-L F1
provides fine-grained credit through their longest common subsequence. In
contrast, a reference is valid only when it identifies the correct source.
These skill-specific signals jointly supervise argument filling and source tracing.

\subsection{Two-Stage GRPO Training}
\label{sec:grpo}

We optimize \methodname{} with Group Relative Policy Optimization
(GRPO)~\cite{grpo} in two stages: Stage~1 trains the workflow planner with the FWG, while Stage~2 trains the tool-call generator to jointly perform tool selection and explicit dataflow modeling.

\paragraph{Stage Objectives.}
For each stage $s\in\{1,2\}$, we define the format reward as
$R_{\mathrm{fmt}}^{(s)}
=\mathds{1}\!\left[\text{the format is valid}\right]$.
The correctness reward for Stage~2 is defined as
\begin{equation}
R_{\mathrm{corr}}^{(2)}
= 2\rho \cdot
\frac{
\lambda_{\mathrm{match}} r_{\mathrm{match}}
+ \lambda_{\mathrm{value}} r_{\mathrm{value}}
}{S_{\max}} -\rho,
\end{equation}
where $\lambda_{\mathrm{match}}$ and $\lambda_{\mathrm{value}}$ weight the two components, and $S_{\max}$ is the maximum attainable weighted score. The total rewards for the two stages are
\begin{equation}
R^{(1)} = R_{\mathrm{fmt}}^{(1)} + R_{\mathrm{func}}^{(1)},
\qquad
R^{(2)} = R_{\mathrm{fmt}}^{(2)} + R_{\mathrm{corr}}^{(2)}.
\end{equation}

\paragraph{Workflow Perturbation for Tool-Call Generation.}
Training on historical workflows focuses the tool-call generator on tool invocation
but may make it overly reliant on perfect workflow inputs, allowing planner
errors to propagate at inference time. We therefore perturb the input workflow with
probability $\epsilon_{\mathrm{pert}}$ while computing the reward against the ground-truth tool calls. This strategy trains the generator to use the query to correct minor planner errors before selecting tools under the resulting constraints. The perturbation operations are detailed in Appendix~\ref{app:details}.

\section{Experiments}
We evaluate \methodname{} by answering the following four research questions:
\begin{itemize}[nosep]
    \item \textbf{RQ1:} Does \methodname{} outperform existing baselines on ID
    benchmarks and generalize to OOD benchmarks with unseen tool sets?
    \item \textbf{RQ2:} Does \methodname{} mitigate the three key limitations of existing
    tool-planning methods?
    \item \textbf{RQ3:} How sensitive is \methodname{} to the number of functional clusters?
    \item \textbf{RQ4:} How does each component of \methodname{} contribute to overall performance?
\end{itemize}
Additional experimental results are provided in Appendix~\ref{app:additional_results}.

\subsection{Experimental Setup}
\label{sec:setup}

\paragraph{Datasets and Evaluation Benchmarks.}
We train \methodname{} on two datasets~\cite{taskbench} for both FWG construction and GRPO training. HuggingFace covers AI model composition tasks, whereas Multimedia targets media processing. We use their held-out test splits for ID evaluation. We also adopt three benchmarks whose tool sets are disjoint from the training tool sets for OOD evaluation:
DailyLifeAPIs~\cite{taskbench} covers everyday API-use tasks;
ToolAlpaca~\cite{toolalpaca} contains simulated APIs across service categories; and Seal-Tools~\cite{sealtools} features challenging multi-tool tasks, including nested tool calls. Dataset statistics are provided in Appendix~\ref{app:exp_details}.

\begin{table*}[!t]
\centering
{%
\small
\begin{tabular}{@{}l@{\hspace{6.85pt}}l*{15}{@{\hspace{6.85pt}}r}@{}}
\toprule
\multirow{2}{*}{\textbf{LLM}} & \multirow{2}{*}{\textbf{Method}}
& \multicolumn{3}{c}{\makebox[0pt][c]{\textbf{HuggingFace} (ID)}}
& \multicolumn{3}{c}{\makebox[0pt][c]{\textbf{Multimedia} (ID)}}
& \multicolumn{3}{c}{\makebox[0pt][c]{\textbf{DailyLifeAPIs} (OOD)}}
& \multicolumn{3}{c}{\makebox[0pt][c]{\textbf{Seal-Tools} (OOD)}}
& \multicolumn{3}{c}{\makebox[0pt][c]{\textbf{ToolAlpaca} (OOD)}} \\
\cmidrule(lr){3-5}\cmidrule(lr){6-8}\cmidrule(lr){9-11}\cmidrule(lr){12-14}\cmidrule(lr){15-17}
& & \makebox[20.5pt][c]{Acc$\uparrow$}
  & \makebox[20.5pt][c]{$n$-F1$\uparrow$}
  & \makebox[20.5pt][c]{$l$-F1$\uparrow$}
  & \makebox[20.5pt][c]{Acc$\uparrow$}
  & \makebox[20.5pt][c]{$n$-F1$\uparrow$}
  & \makebox[20.5pt][c]{$l$-F1$\uparrow$}
  & \makebox[20.5pt][c]{Acc$\uparrow$}
  & \makebox[20.5pt][c]{$n$-F1$\uparrow$}
  & \makebox[20.5pt][c]{$l$-F1$\uparrow$}
  & \makebox[20.5pt][c]{Acc$\uparrow$}
  & \makebox[20.5pt][c]{$n$-F1$\uparrow$}
  & \makebox[20.5pt][c]{$l$-F1$\uparrow$}
  & \makebox[20.5pt][c]{Acc$\uparrow$}
  & \makebox[20.5pt][c]{$n$-F1$\uparrow$}
  & \makebox[20.5pt][c]{$l$-F1$\uparrow$} \\
\midrule
\multirow{6}{*}{Qwen}
& ToolNet
  & 46.67 & 88.72 & 43.52
  & 39.77 & 83.77 & 49.33
  & 41.88 & 89.96 & 24.77
  & 3.94  & 63.44 & 2.41
  & 26.11 & 30.41 & 50.97 \\
& DFSDT
  & 48.02 & 89.07 & 44.49
  & 40.93 & 88.61 & 49.02
  & 51.38 & 90.99 & 22.43
  & 5.02  & 87.63 & 13.99
  & 28.89 & 32.92 & 59.53 \\
& ToolPlanner
  & 49.73 & 85.08 & 45.65
  & 49.41 & 86.45 & 58.86
  & 44.19 & 85.01 & 28.63
  & 30.65 & 83.46 & 25.66
  & 27.61 & 37.56 & 60.55 \\
& GTool
  & 66.28 & 93.38 & 80.89
  & 73.31 & 96.97 & 88.65
  & 61.93 & 95.84 & 75.31
  & 41.32 & 78.84 & 49.36
  & 28.08 & 68.51 & 60.43 \\
& ToolRL
  & \underline{75.68} & \underline{97.84} & \underline{87.28}
  & \underline{78.78} & \underline{98.95} & \underline{92.64}
  & \underline{65.16} & \underline{97.44} & \underline{79.42}
  & \underline{47.67} & \textbf{98.71} & \underline{59.07}
  & \underline{42.08} & \underline{74.16} & \underline{68.66} \\
& \methodname{} (ours)
  & \textbf{76.75} & \textbf{98.25} & \textbf{87.33}
  & \textbf{78.88} & \textbf{99.00} & \textbf{92.69}
  & \textbf{66.19} & \textbf{97.49} & \textbf{79.47}
  & \textbf{56.63} & \underline{98.37} & \textbf{63.47}
  & \textbf{44.44} & \textbf{82.00} & \textbf{68.72} \\
\midrule
\multirow{6}{*}{Llama}
& ToolNet
  & 42.89 & 82.67 & 51.18
  & 40.84 & 86.49 & 46.19
  & 45.19 & 84.95 & 58.04
  & 0.54  & 92.61 & 3.42
  & 27.56 & 54.07 & 55.96 \\
& DFSDT
  & 44.13 & 89.55 & 51.55
  & 39.49 & 90.27 & 46.25
  & 45.45 & 89.21 & 57.96
  & 1.25  & 85.70 & 4.64
  & 25.90 & 54.62 & 55.05 \\
& ToolPlanner
  & 50.84 & 84.96 & 64.91
  & 50.75 & 83.78 & 61.57
  & 21.83 & 71.58 & 57.17
  & 11.29 & 73.23 & 17.46
  & 20.12 & 54.76 & 58.42 \\
& GTool
  & 68.34 & 94.46 & 82.52
  & 73.66 & 97.16 & 89.28
  & 48.22 & 93.22 & 68.25
  & 43.37 & 92.49 & 54.83
  & 30.61 & \underline{70.89} & 61.37 \\
& ToolRL
  & \underline{76.07} & \underline{98.53} & \underline{87.85}
  & \underline{78.88} & \underline{99.18} & \underline{93.32}
  & \underline{64.61} & \underline{97.71} & \underline{77.58}
  & \underline{53.41} & \underline{96.38} & \underline{61.56}
  & \underline{35.78} & 68.32 & \underline{64.09} \\
& \methodname{} (ours)
  & \textbf{77.44} & \textbf{98.71} & \textbf{87.90}
  & \textbf{80.38} & \textbf{99.24} & \textbf{93.37}
  & \textbf{69.30} & \textbf{98.06} & \textbf{81.67}
  & \textbf{56.63} & \textbf{98.68} & \textbf{64.80}
  & \textbf{40.68} & \textbf{82.61} & \textbf{64.21} \\
\bottomrule
\end{tabular}
}
\caption{Tool-planning performance on ID and OOD benchmarks.
Best results are bold; second-best results are underlined.}
\label{tab:main}
\end{table*}

\paragraph{Baselines and Models.}
We compare \methodname{} against five tool-planning baselines.
\textbf{Tool-Planner}~\cite{toolplanner} groups similar APIs into toolkits for
intra-toolkit substitution and cross-toolkit replanning. \textbf{ToolNet}~\cite{toolnet} and \textbf{GTool}~\cite{gtool} construct tool
dependency graphs for graph traversal and tuning-based graph reasoning,
respectively.
\textbf{DFSDT}~\cite{toolllm} performs training-free depth-first tree search,
while \textbf{ToolRL}~\cite{toolrl} trains tool-use policies with reinforcement
learning and fine-grained rewards.
We evaluate all methods with two open-source LLM backbones:
Qwen2.5-7B-Instruct~\cite{Qwen2} and
Llama-3.1-8B-Instruct~\cite{llama}.

\paragraph{Metrics.}
We evaluate the planning performance with three metrics. 
\textbf{Overall Accuracy} (Acc) measures the proportion of plans with correct
tool calls and arguments under rule-based and LLM-as-a-Judge verification
(Appendix~\ref{app:exp_details}). 
\textbf{Node F1} ($n$-F1) computes F1 between the predicted and target tool
sets, 
while \textbf{Link F1} ($l$-F1) computes F1 between their dependency-link
sets~\cite{gtool}. 
To evaluate argument-source tracing, we additionally report
\textbf{Source Error Rate (SER)}, which measures the percentage of incorrectly identified argument sources. 
All metric values are reported on a 0--100 scale.

\paragraph{Implementation Details.}
We construct the FWG with $L{=}30$ functional clusters. We set the reward scale to $\rho{=}3$, with reward weights $\lambda_{\mathrm{match}}{=}1$ and $\lambda_{\mathrm{value}}{=}2$. We train the workflow planner and tool-call generator for 20 and 10 epochs, respectively, and apply workflow perturbation with $\epsilon_{\mathrm{pert}}{=}0.2$. All experiments were conducted on an NVIDIA H200 GPU, with details provided in Appendix~\ref{app:exp_details}.

\subsection{Main Results}
To answer RQ1, we show the results of comparing \methodname{} with five baselines in terms of tool-planning performance in Table~\ref{tab:main}. On the ID benchmarks, \methodname{} achieves the highest Acc with both backbones. With Llama, it outperforms the strongest baseline by 1.37 points and 1.50 points on HuggingFace and Multimedia, respectively, with a similar trend observed for Qwen. It also achieves the best $n$-F1 and $l$-F1 scores in all ID settings, indicating improvements in both tool selection and dependency prediction. 

\methodname{} exhibits more pronounced advantages in OOD evaluations involving unseen tool sets. With Llama, it improves Acc over the strongest baseline by 4.69, 3.22, and 4.90 points on DailyLifeAPIs, Seal-Tools, and ToolAlpaca, respectively. The corresponding $n$-F1 and $l$-F1 results further indicate that these gains extend to tool selection and argument-dependency prediction. Overall, these results demonstrate strong generalization to unseen tool sets.

\subsection{Further Analysis}
\label{sec:rq2}

To answer RQ2, we examine whether \methodname{} addresses three key limitations of existing methods: limited experience available for individual tools, the lack of a global workflow view, and unreliable argument-level dataflow. All experiments in this subsection use Llama-3.1-8B-Instruct, and similar trends are observed with Qwen.

\paragraph{Cross-Tool Experience Sharing.}
We examine whether cross-tool experience sharing through the FWG improves planning for tools with limited historical usage. For each test instance, we count the occurrences of every required tool in the training trajectories and use the minimum count as the instance-level exposure. We then partition the instances into rare, moderate, and frequent groups in a 20/60/20 ratio. We compare \methodname{} with its tool-graph variant, which retains trajectory-derived transitions at the tool level, as well as GTool, a graph-based planning baseline.

\begin{figure}[!t]
\centering
\includegraphics[width=\columnwidth]{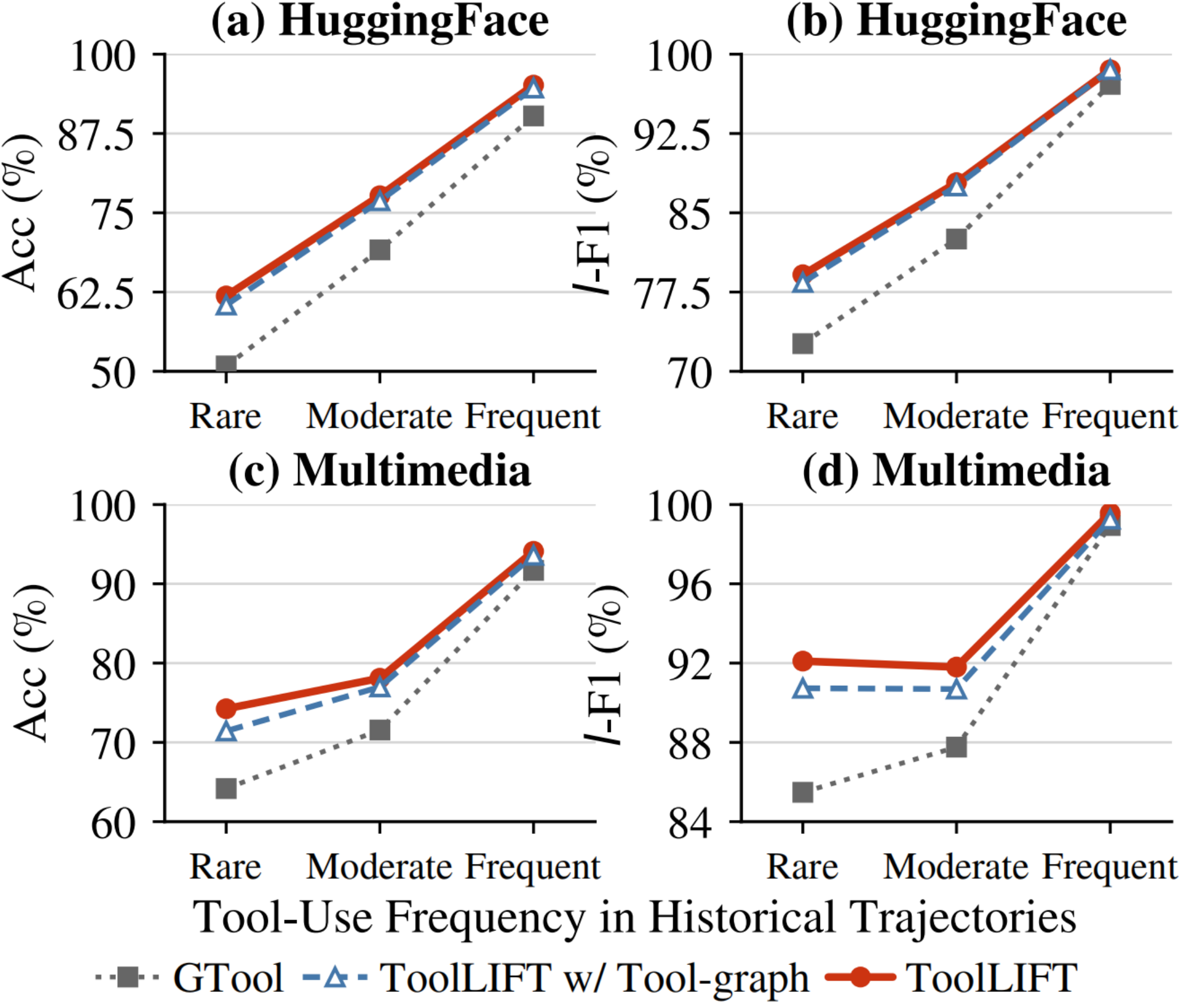}
\caption{Performance across historical tool-use frequencies on HuggingFace and Multimedia.}
\label{fig:experience_sharing}
\end{figure}

Figure~\ref{fig:experience_sharing} shows that \methodname{} consistently outperforms both GTool and its Tool-graph variant. The largest gains over the Tool-graph variant occur in the rare group: Acc and $l$-F1 increase by 1.44 and 0.73 points on HuggingFace, respectively, and by 2.81 and 1.37 points on Multimedia. The smaller gaps in the frequent group indicate that tool-level transitions become more reliable with sufficient observations, whereas trajectory lifting is particularly beneficial when tool-level evidence is sparse.

\paragraph{Global Planning across Tool-Chain Lengths.}
We next examine whether \methodname{} improves awareness of global workflow structure. Longer tool chains require coordinating more interdependent tool calls, making chain length a practical indicator of global planning complexity. We therefore group test instances into short (1--2 calls), medium (3--4 calls), and long ($\geq 5$ calls) chains, and compare \methodname{} with ToolNet and ToolRL.

\begin{figure}[!tbp]
\centering
\includegraphics[width=\columnwidth]{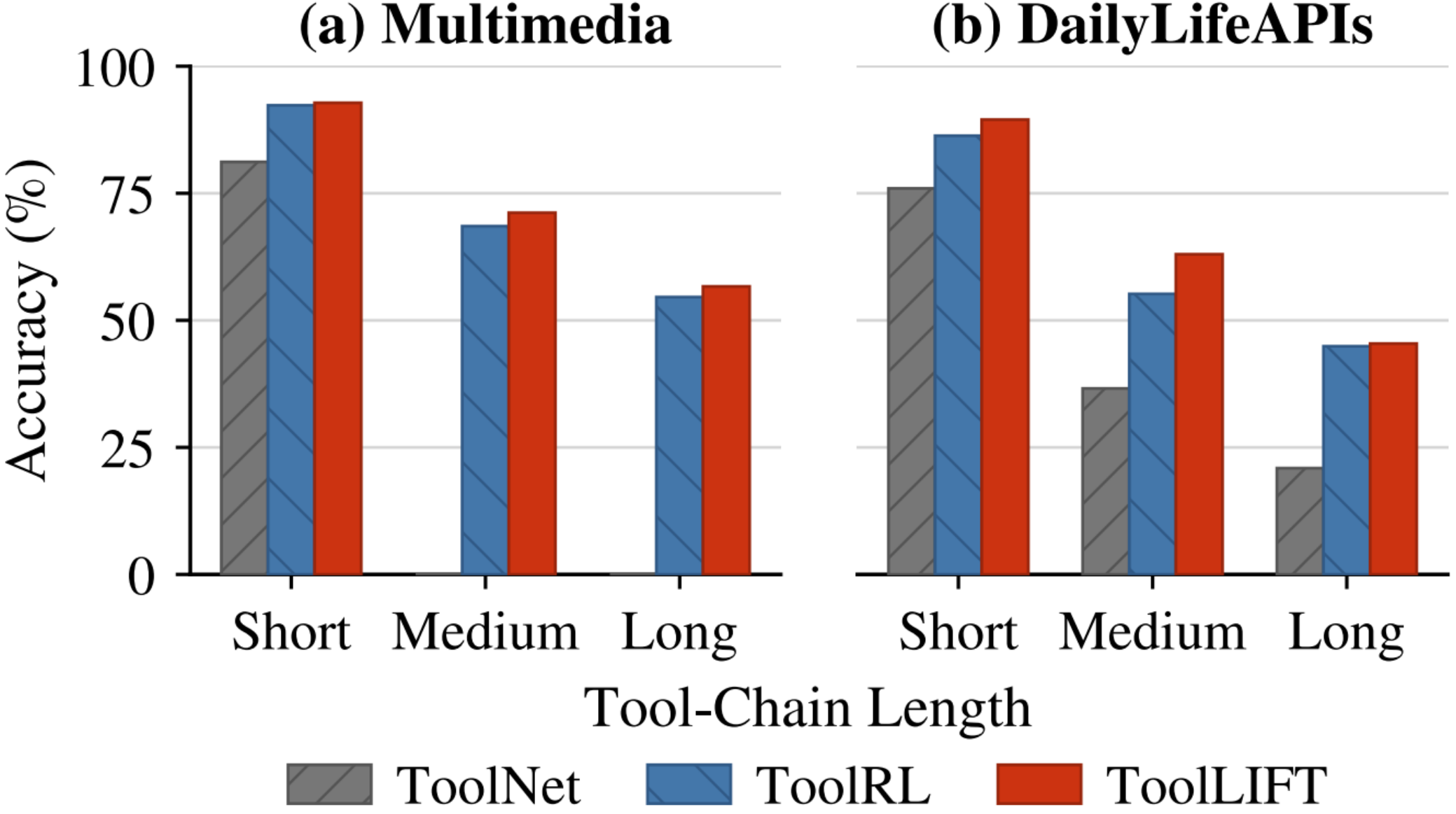}
\caption{Accuracy by tool-chain length on Multimedia (ID) and DailyLifeAPIs (OOD).}
\label{fig:chain_length}
\end{figure}

Figure~\ref{fig:chain_length} shows that \methodname{} outperforms both baselines, with the largest margins on medium chains. Short chains require limited global coordination and therefore leave less room for improvement, whereas medium-length chains benefit more clearly from establishing the complete workflow first. On long chains, errors in tool selection and argument prediction are more likely to accumulate across calls, partially offsetting the benefit of global workflow planning.

\paragraph{Argument Information Source Tracing.}
We examine whether \methodname{} helps trace argument-level dataflow. We compare it with a \methodname{} variant that uses $r_{\mathrm{val}}^{\mathrm{EM}}$ to replace the dependency-aware argument reward with exact-match rewards, thereby removing the explicit learning signal.

\begin{table}[t]
\centering
\small
\setlength{\tabcolsep}{3.5pt}
\begin{tabular}{lccccc}
\toprule
\textbf{Method} & \textbf{HF} & \textbf{MM} & \textbf{Daily} & \textbf{Seal} & \textbf{TA} \\
\midrule
\methodname{} w/ $r_{\mathrm{val}}^{\mathrm{EM}}$ & 17.53 & 8.30 & 10.15 & 15.48 & 15.04 \\
\methodname{} & 14.11 & 6.77 & 5.54 & 9.81 & 12.14 \\
\bottomrule
\end{tabular}
\caption{Source Error Rate (SER, $\downarrow$) across five benchmarks. HF, MM, Daily, Seal and TA denote HuggingFace, Multimedia, DailyLifeAPIs, Seal-Tools and ToolAlpaca, respectively.}
\label{tab:source_tracing}
\end{table}

Table~\ref{tab:source_tracing} shows that \methodname{} reduces source errors on all datasets. This suggests that explicitly learning dataflow helps the generated plan preserve argument-source dependencies.

\paragraph{Sensitivity to the Number of Functional Clusters.}
\label{sec:rq3}
To answer RQ3, we analyze the sensitivity of FWG construction to $L$ by varying $L$ from 10 to 50 in increments of 10.

\begin{figure}[!tbp]
\centering
\includegraphics[width=\columnwidth]{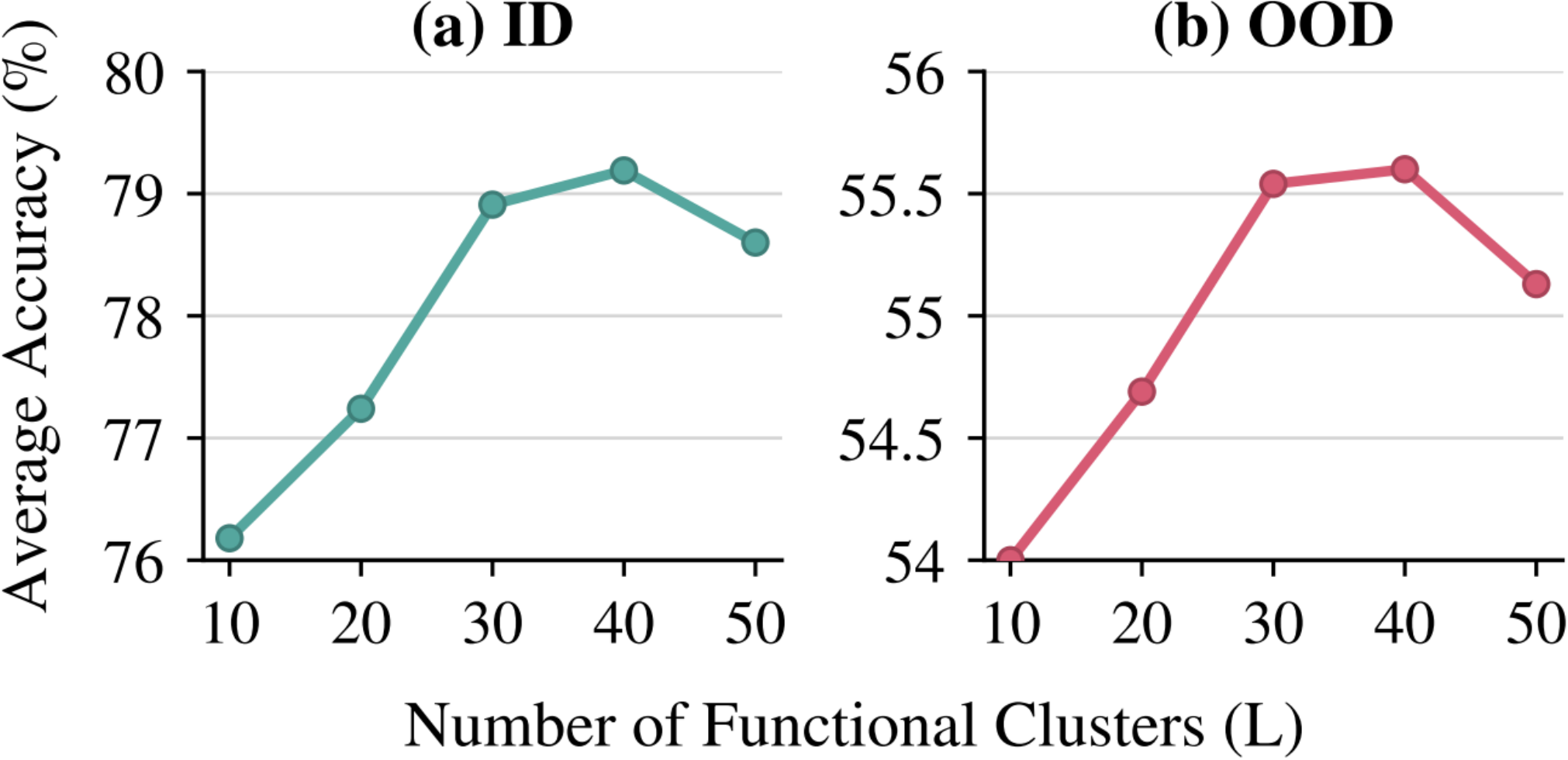}
\caption{Average accuracy under different numbers of functional clusters on ID and OOD benchmarks.} %
\label{fig:cluster_sensitivity}
\end{figure}

Figure~\ref{fig:cluster_sensitivity} shows that average accuracy increases as $L$ grows from 10 to 40, before declining at $L{=}50$. The silhouette-selected $L{=}30$ is already near-optimal, only 0.28 and 0.06 points below the ID and OOD peaks, respectively. This result supports the silhouette coefficient as an effective criterion for selecting $L$. Moreover, it shows that too few clusters merge tools with different functions, making transitions less precise, whereas too many clusters separate functionally similar tools and reduce the trajectory evidence shared within each function. A qualitative analysis of the resulting functional clusters is provided in Appendix~\ref{app:additional_results}.

\subsection{Ablation Study}
\label{sec:rq4}

\begin{table}[t]
\centering
{%
\small
\begin{tabular}{@{}l*{6}{@{\hspace{3.85pt}}r}@{}}
\toprule
\multirow{2}{*}{\textbf{Variant}}
& \multicolumn{3}{c}{\makebox[0pt][c]{\textbf{HF (ID)}}}
& \multicolumn{3}{c}{\makebox[0pt][c]{\textbf{Daily (OOD)}}} \\
\cmidrule(lr){2-4}\cmidrule(lr){5-7}
& \makebox[20.5pt][c]{Acc$\uparrow$}
& \makebox[20.5pt][c]{$n$-F1$\uparrow$}
& \makebox[20.5pt][c]{$l$-F1$\uparrow$}
& \makebox[20.5pt][c]{Acc$\uparrow$}
& \makebox[20.5pt][c]{$n$-F1$\uparrow$}
& \makebox[20.5pt][c]{$l$-F1$\uparrow$} \\
\midrule
Full model
  & \textbf{77.44} & \textbf{98.71} & \textbf{87.90}
  & \textbf{69.30} & \textbf{98.06} & \textbf{81.67} \\
\quad w/o FWG guidance
  & 76.05 & 98.28 & 87.51
  & 60.92 & 97.35 & 70.49 \\
\quad w/o workflow planner
  & 75.58 & 98.10 & 86.89
  & 55.29 & 97.24 & 68.83 \\
\quad w/ $r_{\mathrm{val}}^{\mathrm{EM}}$
  & 75.97 & 98.60 & 87.82
  & 57.96 & 97.64 & 74.08 \\
\quad w/o perturbation
  & 75.43 & 98.31 & 86.81
  & 53.57 & 97.60 & 65.11 \\
\bottomrule
\end{tabular}
}
\caption{Ablation results on HuggingFace (HF) and DailyLifeAPIs (Daily).}
\label{tab:ablation}
\end{table}

To answer RQ4, we report results for four ablation variants in
Table~\ref{tab:ablation}. The \textit{w/o FWG} variant removes the experience
encoded in the FWG while retaining the workflow planner;
\textit{w/o workflow planner} directly predicts tool calls;
the $r_{\mathrm{val}}^{\mathrm{EM}}$ variant replaces the dependency-aware
argument reward with exact-match scoring; and \textit{w/o perturbation} sets
$\epsilon_{\mathrm{pert}}{=}0$, so the generator is trained only with
ground-truth workflows.

Removing the FWG causes a larger Acc drop on DailyLifeAPIs than on HuggingFace, indicating its value for transferring collaboration structures to unseen tool sets. Removing workflow planning also causes larger drops on DailyLifeAPIs, showing that function-level planning provides useful task-level context for tool selection and dependency prediction. Using $r_{\mathrm{val}}^{\mathrm{EM}}$ weakens the learning signal for argument-source identification. Finally, removing workflow perturbation causes the largest OOD degradation and also lowers ID performance, showing its role in preventing minor workflow errors from propagating to the tool-call plan.

\section{Conclusion}
In this paper, we introduce \methodname{}, which lifts tool-specific trajectories into FWG for
generalizable tool planning. Our key observation is that analogous tasks often share
common function-level workflow structures despite using different concrete tools. Based on this
observation, \methodname{} constructs the FWG through trajectory lifting,
decouples global workflow planning from concrete tool selection, and explicitly
models argument-level dataflow for source-traceable tool calls. Experiments on
two ID and three OOD benchmarks demonstrate consistent performance gains, with
particularly strong generalization to unseen tool sets. One limitation is our
assumption that each argument has a single information source. Extending the
framework to support arguments derived jointly from multiple context items or
tool outputs remains an important direction for future work.

\bibliography{aaai2027}
\clearpage
\appendix
\section{Method Details}
\label{app:details}
This section provides the algorithmic, serialization, and prompt details of \methodname{}.
For a ground-truth tool-call plan $\mathcal{P}^{*}=(p_1^{*},\ldots,p_{K^{*}}^{*})$, we obtain its target function-level workflow by applying the tool-to-function lifting:
\begin{equation}
\mathcal{W}_c^{*}=
(\phi(t_1^{*}),\ldots,\phi(t_{K^{*}}^{*})).
\end{equation}

\subsection{Two-Stage Training and Inference}
Algorithm~\ref{alg:two_stage_training} first optimizes the workflow planner with $R^{(1)}$ and then optimizes the tool-call generator with $R^{(2)}$.
During Stage~2 training, the generator receives either the ground-truth workflow or its perturbed version, while its correctness reward is always computed against the ground-truth tool-call plan.
At inference time, the generator receives the workflow predicted by the trained planner.

\begin{algorithm}[!b]
\caption{Two-stage training and inference}
\label{alg:two_stage_training}
\begin{algorithmic}[1]
\STATE \textbf{Input:} Training instances $(q,\mathcal{T}_q,\mathcal{P}^{*})$; mapping $\phi$; FWG $\mathcal{G}_{\mathrm{fwg}}$; workflow-perturbation probability $\epsilon_{\mathrm{pert}}$
\STATE \textbf{Output:} Workflow planner $\pi_{\theta_1}$ and tool-call generator $\pi_{\theta_2}$
\STATE Map each $\mathcal{P}^{*}$ to $\mathcal{W}_c^{*}$ with $\phi$
\FOR{each Stage~1 training batch}
  \STATE Extract the query-specific FWG subgraph $\mathcal{G}_q$ for each instance
  \STATE Build planner prompt $x^{(1)}$ from $q$, $\mathcal{T}_q$, and $\mathcal{G}_q$
  \STATE Sample $G$ workflows $\{\hat{\mathcal{W}}_{c,i}\}_{i=1}^{G}$ from $\pi_{\theta_1}$
  \STATE Score each output with
  $R_i^{(1)}=R_{\mathrm{fmt},i}^{(1)}+R_{\mathrm{plan},i}^{(1)}$
  \STATE Update $\theta_1$ with GRPO using $\{R_i^{(1)}\}_{i=1}^{G}$
\ENDFOR
\FOR{each Stage~2 training batch}
  \FOR{each instance in the batch}
    \STATE Sample $b\sim\mathrm{Bernoulli}(\epsilon_{\mathrm{pert}})$
    \IF{$b=1$}
      \STATE $\widetilde{\mathcal{W}}_c\leftarrow\operatorname{PerturbWorkflow}(\mathcal{W}_c^{*})$
    \ELSE
      \STATE $\widetilde{\mathcal{W}}_c\leftarrow\mathcal{W}_c^{*}$
    \ENDIF
    \STATE Build $x^{(2)}$ from $q$, $\mathcal{T}_q$, and $\widetilde{\mathcal{W}}_c$
  \ENDFOR
  \STATE Sample $G$ tool-call plans $\{\hat{\mathcal{P}}_i\}_{i=1}^{G}$ from $\pi_{\theta_2}$
  \STATE Score each output against $\mathcal{P}^{*}$ with
  $R_i^{(2)}=R_{\mathrm{fmt},i}^{(2)}+R_{\mathrm{corr},i}^{(2)}$
  \STATE Update $\theta_2$ with GRPO using $\{R_i^{(2)}\}_{i=1}^{G}$
\ENDFOR
\STATE \textbf{Inference:} predict $\hat{\mathcal{W}}_c$ with $\pi_{\theta_1}$, then predict $\hat{\mathcal{P}}$ with $\pi_{\theta_2}$ conditioned on $\hat{\mathcal{W}}_c$
\end{algorithmic}
\end{algorithm}

\subsection{Workflow Perturbation}
Algorithm~\ref{alg:workflow_perturbation} applies one local edit to the ground-truth function-level workflow.
We sample deletion, insertion, and replacement with probabilities $0.45$, $0.35$, and $0.20$, respectively.
Positions are sampled uniformly, and the function used for insertion or replacement is sampled uniformly from $\mathcal{C}$, excluding the original function in the replacement case.
The perturbation changes the workflow guidance but not the ground-truth tool-call plan used by the Stage~2 reward.

\begin{algorithm}[!b]
\caption{Workflow perturbation}
\label{alg:workflow_perturbation}
\begin{algorithmic}[1]
\STATE \textbf{Input:} Nonempty ground-truth function-level workflow $\mathcal{W}_c^{*}=(c_1,\ldots,c_K)$; function set $\mathcal{C}$
\STATE \textbf{Output:} Perturbed workflow $\widetilde{\mathcal{W}}_c$
\STATE Sample $o\sim\mathrm{Categorical}(0.45,0.35,0.20)$ over $\{\textsc{Delete},\textsc{Insert},\textsc{Replace}\}$
\STATE $\widetilde{\mathcal{W}}_c\leftarrow\mathcal{W}_c^{*}$
\IF{$o=\textsc{Delete}$}
  \STATE Sample $j\sim\mathrm{Uniform}(\{1,\ldots,K\})$ and delete $c_j$
\ELSIF{$o=\textsc{Insert}$}
  \STATE Sample $j\sim\mathrm{Uniform}(\{0,\ldots,K\})$ and $c'\sim\mathrm{Uniform}(\mathcal{C})$
  \STATE Insert $c'$ after position $j$
\ELSE
  \STATE Sample $j\sim\mathrm{Uniform}(\{1,\ldots,K\})$ and $c'\sim\mathrm{Uniform}(\mathcal{C}\setminus\{c_j\})$
  \STATE Replace $c_j$ with $c'$
\ENDIF
\RETURN $\widetilde{\mathcal{W}}_c$
\end{algorithmic}
\end{algorithm}

\subsection{GRPO Optimization}
We apply the same GRPO update independently in both stages.
For prompt $x^{(s)}$ at stage $s\in\{1,2\}$, the old policy samples $G$ outputs, each of which is scored with the corresponding stage reward $R_i^{(s)}$.
We apply reference-policy regularization to the sampled sequence score:
\begin{equation}
\widetilde{R}_i^{(s)}=R_i^{(s)}-
\beta\sum_{t=1}^{|y_i|}
\left(\log\pi_{\theta_{s,\mathrm{old}}}(y_{i,t})-
\log\pi_{\mathrm{ref}}(y_{i,t})\right),
\end{equation}
Both probabilities condition on $x^{(s)}$ and $y_{i,<t}$.
We set $\beta=10^{-3}$ and do not add a separate KL term to the actor loss.
GRPO then forms the group-relative advantage
\begin{equation}
\hat{A}_i^{(s)}=
\frac{\widetilde{R}_i^{(s)}-
\operatorname{mean}_{j}(\widetilde{R}_j^{(s)})}
{\operatorname{std}_{j}(\widetilde{R}_j^{(s)})+\epsilon_a}.
\end{equation}
Let
\begin{equation}
\begin{aligned}
\eta_{i,t}(\theta_s)
&=
\frac{
\pi_{\theta_s}(y_{i,t}\mid x^{(s)},y_{i,<t})
}{
\pi_{\theta_{s,\mathrm{old}}}(y_{i,t}\mid x^{(s)},y_{i,<t})
},\\
\bar{\eta}_{i,t}
&=
\operatorname{clip}\!\left(
\eta_{i,t}(\theta_s),1-\epsilon,1+\epsilon
\right).
\end{aligned}
\end{equation}
The stage policy is updated with
\begin{equation}
\mathcal{J}^{(s)}(\theta_s)=
\mathbb{E}\!\left[
\frac{1}{G}\sum_{i=1}^{G}\frac{1}{|y_i|}\sum_{t=1}^{|y_i|}
\ell_{i,t}^{(s)}(\theta_s)\right],
\end{equation}
where
\begin{equation}
\ell_{i,t}^{(s)}(\theta_s)=
\min\!\left(\eta_{i,t}\hat{A}_i^{(s)},
\bar{\eta}_{i,t}\hat{A}_i^{(s)}\right).
\end{equation}
This critic-free update uses relative comparisons among structured outputs for the same prompt.

\subsection{Textual Serialization of FWG Transitions}
The FWG stores transition probabilities between functional clusters.
For each function in the query-specific subgraph, we rank its outgoing neighbors by transition probability and serialize the resulting transitions as text.

\begin{promptbox}{Textual Serialization of FWG transitions}
\textbf{Serialization Template}
\begin{quote}
\ttfamily
Cluster Transition Patterns (rank next cluster from historical workflows):\\
Use these patterns as soft guidance.
They reflect common workflows but do not constrain your plan.\\
cluster \{source\_cluster\}: cluster \{rank\_1\}, cluster \{rank\_2\}, cluster \{rank\_3\}\\
...
\end{quote}

\textbf{Example}
\begin{quote}
\ttfamily
cluster 0: cluster 5, cluster 0, cluster 18\\
cluster 1: cluster 5, cluster 18, cluster 1\\
cluster 5: cluster 18, cluster 5, cluster 0\\
cluster 18: cluster 1, cluster 5, cluster 18
\end{quote}
\end{promptbox}

\subsection{Prompt Details}
\label{app:prompts}
This subsection provides the prompts and output templates.

\paragraph{Tool Description Decomposition Prompt.}
This prompt decomposes each tool schema into a domain description and an abstract functional description. We execute this prompt using Deepseek-v4\cite{deepseekv4}.

\begin{promptbox}{Tool Description Decomposition Prompt}
You are an expert system specializing in analyzing and abstracting the underlying logic of APIs and tools.
Your task is to read the given tool descriptions and extract two core features for high-dimensional tool clustering.

\textbf{Extraction Tasks}
\begin{enumerate}
    \item \textbf{Domain:} What specific business, industry, or vertical domain does this tool serve?
    Summarize it in one short sentence or phrase.
    \item \textbf{Abstract Function:} Completely ignore domain-specific nouns such as stocks, weather, flights, or medical records.
    Reduce the tool's core action to the most fundamental computer-science, logic, data, or resource operation.
    Summarize it in one sentence.
\end{enumerate}

\textbf{Output Format}
\begin{quote}
\ttfamily
Domain: [One-sentence description]\\
Abstract Function: [One-sentence description]
\end{quote}

\textbf{Examples}

\begin{quote}
\ttfamily
Input:\\
Tool Name: get\_stock\_price\\
Tool Description: Input the company's stock ticker, e.g., AAPL, to get the current real-time trading price from the NASDAQ exchange.\\
Output:\\
Domain: Financial trading and stock market.\\
Abstract Function: Retrieve and return the real-time numerical state of an entity based on its unique identifier.
\end{quote}

\begin{quote}
\ttfamily
Input:\\
Tool Name: book\_flight\_ticket\\
Tool Description: Input departure, destination, time, and passenger info to lock a seat and generate a flight ticket order in the airline system.\\
Output:\\
Domain: Aviation travel and ticket booking.\\
Abstract Function: Receive multi-dimensional attribute parameters, verify quotas in the system, and create a resource occupation record.
\end{quote}
\end{promptbox}

\paragraph{Function-Level Workflow Planning Prompt.}
This prompt asks the model to generate a complete function-level workflow from the user query, candidate functions, and the serialized query-specific FWG subgraph.

\begin{promptbox}{Function-Level Workflow Planning Prompt}
You are a high-level Workflow Planner assistant.
Your objective is to understand the user's task and tool clusters, and then identify the logical cluster workflow required to solve the task.

\textbf{Available Tool Clusters}

\texttt{\{AVAILABLE\_TOOL\_CLUSTERS\}}

\textbf{Cluster Transition Patterns (rank next cluster from historical workflows):}

Use these patterns as soft guidance.
They reflect common workflows but do not constrain your plan.

\texttt{\{CLUSTER\_TRANSITION\_PATTERNS\}}

\textbf{Output Format (for example)}

\begin{quote}
\ttfamily
<think> Your thoughts on identifying the cluster workflow</think>\\
<plan>\\
cluster 1 -> cluster 2\\
</plan>
\end{quote}
\end{promptbox}

\paragraph{Tool-Call Generation Prompt.}
This prompt asks the model to verify the planned workflow, select concrete tools, and generate a tool-call plan with explicit argument sources.

\begin{promptbox}{Tool-Call Generation Prompt}
You are a precise tool-executing assistant.
For each request, you MUST first reason in \texttt{\textless think\textgreater...\textless/think\textgreater}: verify the query, check whether the Planned Workflow is correct, choose tools from the listed clusters, and plan argument dependencies.
Only after that, output exactly one JSON array of tool calls.
Never skip the think block or output bare JSON.

\textbf{\#\#\# Available Tools}

\texttt{\{AVAILABLE\_TOOLS\}}

\textbf{\#\#\# Strict Rules}
\begin{enumerate}
    \item Dependencies: If an argument requires the output of another tool being called right now, use the exact format \texttt{(need\_output\_from\_ToolName)}.
    \item Follow the Expected Output Format below exactly.
    \item Before outputting the result JSON, you MUST do some thinking.
    Wrap your thought process in \texttt{\textless think\textgreater\textless/think\textgreater}.
\end{enumerate}

\textbf{\#\#\# Expected Output Format}
\begin{quote}
\ttfamily
\textless think\textgreater\\
Reason about which tool in each cluster of the planned workflow best fits the task, how to fill its arguments, and how data flows between them.\\
\textless/think\textgreater\\
{[}\\
\hspace*{1em}\{\\
\hspace*{2em}"name": "First Tool Name",\\
\hspace*{2em}"arguments": \{\\
\hspace*{3em}"argument\_name": "provided\_value"\\
\hspace*{2em}\}\\
\hspace*{1em}\},\\
\hspace*{1em}\{\\
\hspace*{2em}"name": "Second Tool Name",\\
\hspace*{2em}"arguments": \{\\
\hspace*{3em}"argument\_name": "(need\_output\_from\_First Tool Name)"\\
\hspace*{2em}\}\\
\hspace*{1em}\}\\
{]}
\end{quote}
\end{promptbox}

\FloatBarrier
\section{Experimental Setup}
\label{app:exp_details}
This section describes dataset processing, evaluation metrics, and implementation settings.

\subsection{Datasets and Data Preprocessing}
\begin{table}[!htbp]
\centering
\small
\begin{tabular}{lccc}
\toprule
\textbf{Dataset} & \textbf{\# Samples} & \textbf{\# Tools} & \textbf{Avg.\ Calls} \\
\midrule
HuggingFace & 3{,}000 & 584 & 2.83 \\
Multimedia  & 3{,}000 & 294 & 2.89 \\
\bottomrule
\end{tabular}
\caption{Statistics of the two training datasets.}
\label{tab:data}
\end{table}

We preprocess the original test data before evaluation.
Table~\ref{tab:eval_data_curation} reports the raw test size and the number of instances retained after preprocessing.
For HuggingFace and Multimedia, the raw test size excludes the 3{,}000 training instances used for FWG construction and GRPO training.
We apply two general criteria across the five evaluation datasets.
First, we remove instances whose target arguments require information that cannot be inferred from the user request or preceding tool outputs.
Second, we downsample single-tool instances so that the evaluation emphasizes the multi-tool planning setting studied in this work.
We then convert all retained examples to a unified representation.

\begin{table}[!htbp]
\centering
\small
\begin{tabular}{lrr}
\toprule
\textbf{Dataset} & \textbf{Raw Test} & \textbf{Final Test} \\
\midrule
HuggingFace & 4{,}094 & 3{,}635 \\
Multimedia & 2{,}361 & 2{,}008 \\
DailyLifeAPIs & 4{,}200 & 3{,}866 \\
Seal-Tools & 558 & 558 \\
ToolAlpaca & 4{,}255 & 1{,}143 \\
\bottomrule
\end{tabular}
\caption{Evaluation-set sizes after preprocessing.}
\label{tab:eval_data_curation}
\end{table}

For HuggingFace, Multimedia, and DailyLifeAPIs, which are subsets of TaskBench, we use DeepSeek-V4 to complete missing fields in the tool descriptions.
For Seal-Tools, we retain all multi-tool instances from the original data.
For ToolAlpaca, we apply a source-grounding audit and remove targets whose arguments are supported by neither the user request nor preceding tool outputs.

\subsection{Evaluation Protocol and Metrics}
We compute accuracy with a two-stage evaluation protocol.
The first stage uses a deterministic rule checker to parse the predicted and ground-truth tool calls.
It marks an instance as correct when each ground-truth call is covered by a predicted call with the same tool name and all required arguments match exactly or numerically.
It marks an instance as a hard error when a required tool is missing or a required argument has the wrong source type.
The remaining cases primarily contain argument-value mismatches.
Because semantically equivalent dates, free-text values, and alternative surface forms may differ as strings, we send these soft errors to an LLM verifier using the prompt in Appendix~\ref{app:accuracy_verification_prompt}.
Final accuracy is the fraction of instances accepted by either stage.

For $n$-F1, let $\mathcal{N}_i$ and $\widehat{\mathcal{N}}_i$ denote the ground-truth and predicted tool-name sets for instance $i$.
We compute
\begin{equation}
n\text{-}\mathrm{F1}_i =
\frac{2|\widehat{\mathcal{N}}_i \cap \mathcal{N}_i|}
{|\widehat{\mathcal{N}}_i| + |\mathcal{N}_i|}.
\end{equation}
The reported $n$-F1 is the average over all evaluated instances and measures tool-selection quality.

For $l$-F1, we derive directed dependency links from the reference arguments in each tool-call plan.
Let $\mathcal{L}_i$ and $\widehat{\mathcal{L}}_i$ denote the ground-truth and predicted link sets for instance $i$. Then we have
\begin{equation}
l\text{-}\mathrm{F1}_i =
\frac{2|\widehat{\mathcal{L}}_i \cap \mathcal{L}_i|}
{|\widehat{\mathcal{L}}_i| + |\mathcal{L}_i|},
\end{equation}
and set $\mathrm{l\text{-}F1}_i=1$ when both link sets are empty.
The reported $l$-F1 is the average over all evaluated instances and measures recovery of the cross-call dependency structure.

For argument-source tracing, we report the source error rate (SER) on samples with correct tool selection.
Let $\mathcal{A}$ be the set of required arguments in these samples, and let $\operatorname{src}(a)$ and $\widehat{\operatorname{src}}(a)$ denote the ground-truth and predicted sources of argument $a$.
An argument source is either a direct value from the context or the output of an upstream tool call.
\begin{equation}
\mathrm{SER} =
\frac{\sum_{a\in\mathcal{A}}\indicator
[\widehat{\operatorname{src}}(a)\ne\operatorname{src}(a)]}{|\mathcal{A}|}.
\end{equation}

\paragraph{Accuracy Verification Prompt.}
\label{app:accuracy_verification_prompt}
This prompt is used in the second accuracy-evaluation stage to verify soft-error cases whose argument do not exactly match the ground truth.

\begin{promptbox}{Accuracy Verification Prompt}
You are a strict but fair evaluator for tool-call predictions.
Given a user instruction, ground-truth tool calls, and predicted tool calls, judge whether the prediction can successfully accomplish the user's task.

\textbf{Evaluation Rule}
Check the ground truth against the prediction.
For each ground-truth tool call, find a predicted call with the same tool name and verify that every required ground-truth argument is satisfied.
Ignore extra predicted arguments and extra predicted tool calls.

\textbf{Argument Matching}
\begin{enumerate}
    \item \textbf{Dependency arguments:} if the ground-truth value uses \texttt{(need\_output\_from\_ToolName)}, the prediction must also use a dependency placeholder pointing to the same upstream tool.
    \item \textbf{Direct arguments:} entity names, identifiers, enum values, file paths, Boolean flags, dates, times, and numeric values should preserve the same semantics as the ground truth.
    Free-text arguments may differ in wording but must retain all key information.
\end{enumerate}

\textbf{Output Format}
\begin{quote}
\ttfamily
\{"thought": "[one-sentence explanation]", "correct": true/false\}
\end{quote}
\end{promptbox}

\subsection{FWG Construction Configuration}
Table~\ref{tab:fwg_construction_config} summarizes the key UMAP and $K$-means settings used to construct the FWG.

\begin{table}[!htbp]
\centering
\small
\begin{tabular}{@{}lll@{}}
\toprule
\textbf{Component} & \textbf{Hyperparameter} & \textbf{Value} \\
\midrule
\multirow{5}{*}{UMAP}
& Number of Neighbors & 15 \\
& Number of Components & 32 \\
& Distance Metric & Cosine \\
& Minimum Distance & 0.1 \\
& Random Seed & 42 \\
\midrule
\multirow{5}{*}{$K$-means}
& Number of Clusters & 30 \\
& Initialization & $K$-means++ \\
& Number of Initializations & Auto \\
& Maximum Iterations & 300 \\
& Random Seed & 42 \\
\bottomrule
\end{tabular}
\caption{Key configuration for FWG construction.}
\label{tab:fwg_construction_config}
\end{table}
\FloatBarrier

\subsection{Training Hyperparameters}
Tables~\ref{tab:stage1_training_config} and \ref{tab:stage2_training_config} summarize the main training hyperparameters for workflow planning and tool-call generation.

\begin{table}[!htbp]
\centering
\small
\begin{tabular}{@{}p{0.42\columnwidth}p{0.50\columnwidth}@{}}
\toprule
\textbf{Hyperparameter} & \textbf{Value} \\
\midrule
\multicolumn{2}{@{}l}{\textbf{Data Configuration}} \\
\midrule
Train Batch Size & 512 \\
Validation Batch Size & 128 \\
Max Prompt Length & 2048 \\
Max Response Length & 256 \\
\midrule
\multicolumn{2}{@{}l}{\textbf{Optimization}} \\
\midrule
Algorithm & GRPO \\
Learning Rate & 1e-6 \\
PPO Mini Batch Size & 32 \\
PPO Micro Batch Size & 4 \\
KL Loss Used & False \\
Reward-Side KL Coefficient & 1e-3 \\
Optimizer & 8-bit AdamW \\
\midrule
\multicolumn{2}{@{}l}{\textbf{Rollout Configuration}} \\
\midrule
Rollout Engine & vLLM \\
GPU Memory Utilization & 0.5 \\
Tensor Parallel Size & 1 \\
Number of Rollouts & 4 \\
\midrule
\multicolumn{2}{@{}l}{\textbf{Training}} \\
\midrule
Total Epochs & 20 \\
\bottomrule
\end{tabular}
\caption{Configuration for workflow-planning GRPO.}
\label{tab:stage1_training_config}
\end{table}

\begin{table}[!htbp]
\centering
\small
\begin{tabular}{@{}p{0.42\columnwidth}p{0.50\columnwidth}@{}}
\toprule
\textbf{Hyperparameter} & \textbf{Value} \\
\midrule
\multicolumn{2}{@{}l}{\textbf{Data Configuration}} \\
\midrule
Train Batch Size & 512 \\
Validation Batch Size & 128 \\
Max Prompt Length & 2048 \\
Max Response Length & 1024 \\
\midrule
\multicolumn{2}{@{}l}{\textbf{Optimization}} \\
\midrule
Algorithm & GRPO \\
Learning Rate & 1e-6 \\
PPO Mini Batch Size & 32 \\
PPO Micro Batch Size & 4 \\
KL Loss Used & False \\
Reward-Side KL Coefficient & 1e-3 \\
Optimizer & 8-bit AdamW \\
\midrule
\multicolumn{2}{@{}l}{\textbf{Rollout Configuration}} \\
\midrule
Rollout Engine & vLLM \\
GPU Memory Utilization & 0.5 \\
Tensor Parallel Size & 1 \\
Number of Rollouts & 4 \\
\midrule
\multicolumn{2}{@{}l}{\textbf{Training}} \\
\midrule
Total Epochs & 10 \\
\bottomrule
\end{tabular}
\caption{Configuration for tool-call-generation GRPO.}
\label{tab:stage2_training_config}
\end{table}
\FloatBarrier

\subsection{Inference Settings}
Table~\ref{tab:inference_config} reports the settings used for end-to-end generation.
The workflow planner first predicts a function-level workflow, after which the tool-call generator produces the final plan conditioned on that workflow.

\begin{table}[!htbp]
\centering
\small
\begin{tabular}{@{}p{0.42\columnwidth}p{0.50\columnwidth}@{}}
\toprule
\textbf{Hyperparameter} & \textbf{Value} \\
\midrule
\multicolumn{2}{@{}l}{\textbf{Workflow Planning}} \\
\midrule
Serving Engine & vLLM \\
Max Prompt Length & 2048 \\
Max Generation Tokens & 256 \\
Temperature & 0 \\
Max Model Length & 2304 \\
\midrule
\multicolumn{2}{@{}l}{\textbf{Tool-Call Generation}} \\
\midrule
Serving Engine & vLLM \\
Temperature & 0 \\
Prefix Caching & Enabled \\
\midrule
\multicolumn{2}{@{}l}{\textbf{Runtime Configuration}} \\
\midrule
GPU Memory Utilization & 0.3 \\
Tensor Parallel Size & 1 \\
Max Concurrent Sequences & 1024 (Workflow Planning), 256 (Tool-Call Generation) \\
\bottomrule
\end{tabular}
\caption{Inference configuration for end-to-end generation.}
\label{tab:inference_config}
\end{table}

\subsection{Computing Infrastructure and Runs}
All experiments were conducted on a server running Ubuntu 24.04.2 LTS with two Intel Xeon Platinum 8558 CPUs.
Each experiment used one NVIDIA H200 GPU with 141~GiB of memory.
The software environment comprised Python 3.12.3, PyTorch 2.4.0 with CUDA 12.1, Transformers 4.47.1, vLLM 0.6.3, VERL 0.1, scikit-learn 1.8.0, UMAP 0.5.11, bitsandbytes 0.49.2, and FlashAttention 2.8.3.
Unless otherwise stated, each reported result was obtained from one training run followed by one deterministic evaluation run.

\FloatBarrier
\section{Additional Experimental Results and Analyses}
\label{app:additional_results}
This section reports supplementary results and qualitative analyses that support the main analysis.

\subsection{Sensitivity of Workflow Perturbation Probability}
\label{app:workflow_perturbation}
\begin{figure}[tbp]
  \centering
  \includegraphics[width=\linewidth]{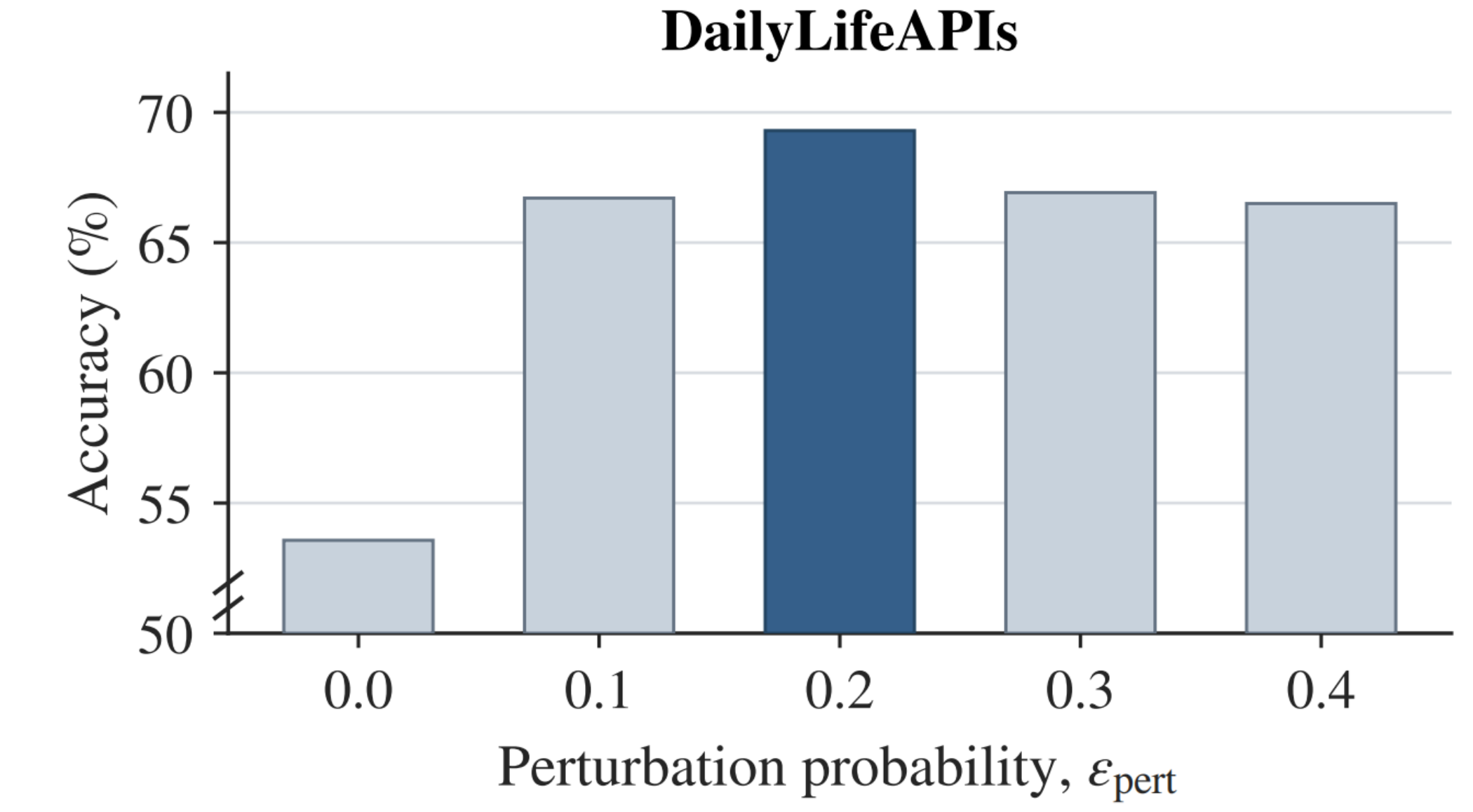}
  \caption{Sensitivity to the workflow-perturbation probability $\epsilon_{\mathrm{pert}}$ on DailyLifeAPIs.}
  \label{fig:workflow_perturbation_dailylifeapis}
\end{figure}

\begin{figure}[tbp]
  \centering
  \includegraphics[width=\linewidth]{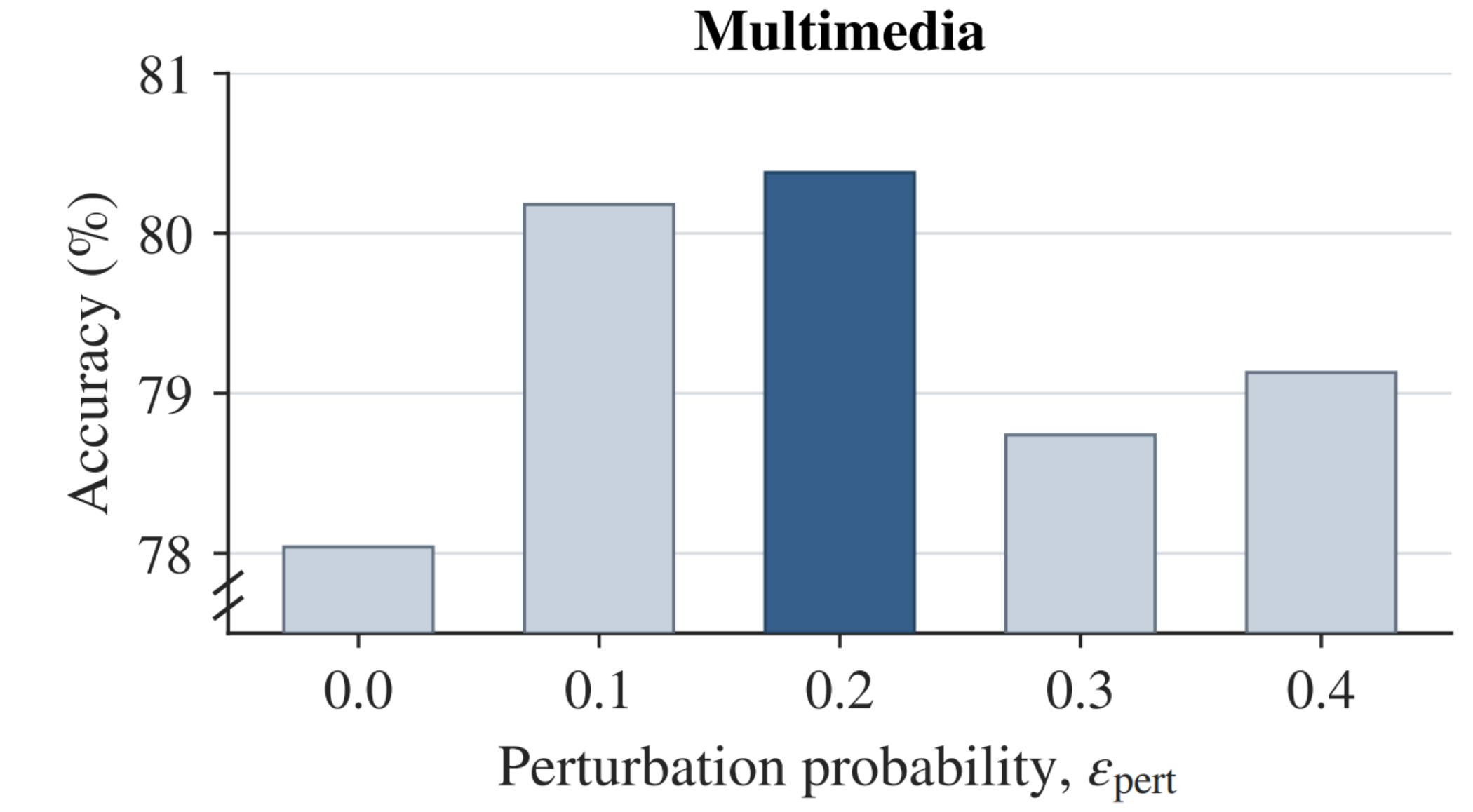}
  \caption{Sensitivity to the workflow-perturbation probability $\epsilon_{\mathrm{pert}}$ on Multimedia.}
  \label{fig:workflow_perturbation_multimedia}
\end{figure}

We vary the workflow-perturbation probability $\epsilon_{\mathrm{pert}}$ during tool-call-generator training and report accuracy on DailyLifeAPIs and Multimedia.
As shown in Figures~\ref{fig:workflow_perturbation_dailylifeapis} and \ref{fig:workflow_perturbation_multimedia}, $\epsilon_{\mathrm{pert}}{=}0.2$ performs best on both datasets.
A lower probability provides insufficient exposure to errors, whereas a higher probability overexposes it to perturbed workflows and weakens its adherence to correct workflow guidance.

\subsection{Qualitative Analysis of Functional Clustering}
\label{app:cluster_analysis}

\begin{figure}[t]
    \centering
    \includegraphics[width=\columnwidth]{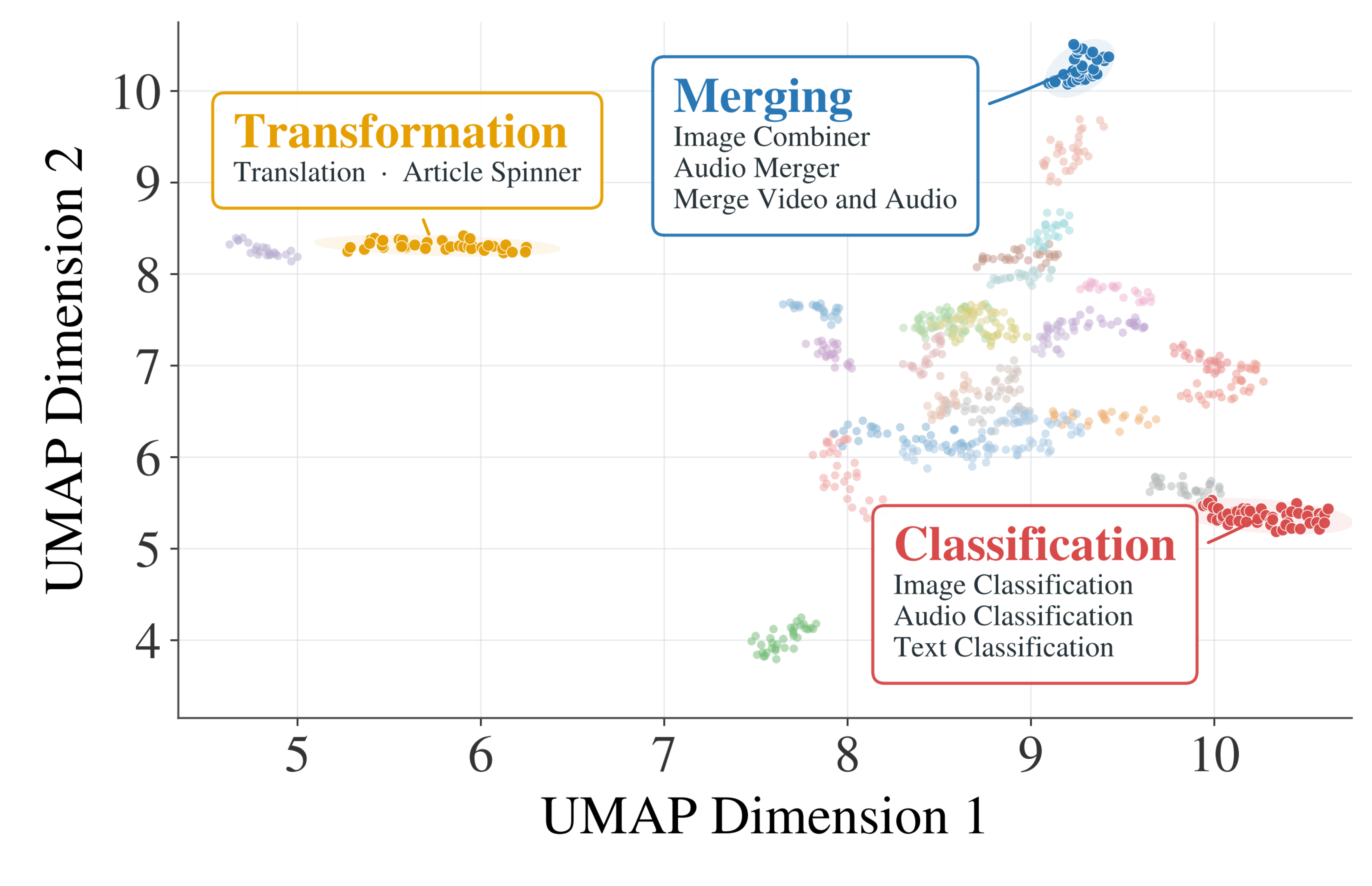}
    \caption{Qualitative visualization of tool functional clustering.
    Each point represents a tool and is colored according to its K-Means
    assignment with $L{=}30$. Three representative functional clusters
    are annotated with example tools.}
    \label{fig:cluster_visualization}
\end{figure}

We qualitatively inspect the functional clusters obtained with the silhouette-selected $L{=}30$. Figure~\ref{fig:cluster_visualization} plots the two-dimensional
UMAP representations of the tools, with each point colored according to its
K-Means cluster assignment. For readability, we annotate three representative
clusters and list several tools from each cluster.

The highlighted clusters exhibit coherent functional abstractions across
different concrete implementations. For example, the
\textit{Classification} cluster groups image, audio, and text classification
tools despite their different input modalities. Similarly, the
\textit{Merging} cluster contains tools that combine image, audio, or
video content. The \textit{Transformation} cluster groups tools that
rewrite textual content through operations such as translation and
paraphrasing. These examples qualitatively indicate that the clustering can
capture shared functional roles beyond individual tool names and modalities.

\subsection{Training Dynamics}
\begin{figure}[!htbp]
  \centering
  \includegraphics[width=\linewidth]{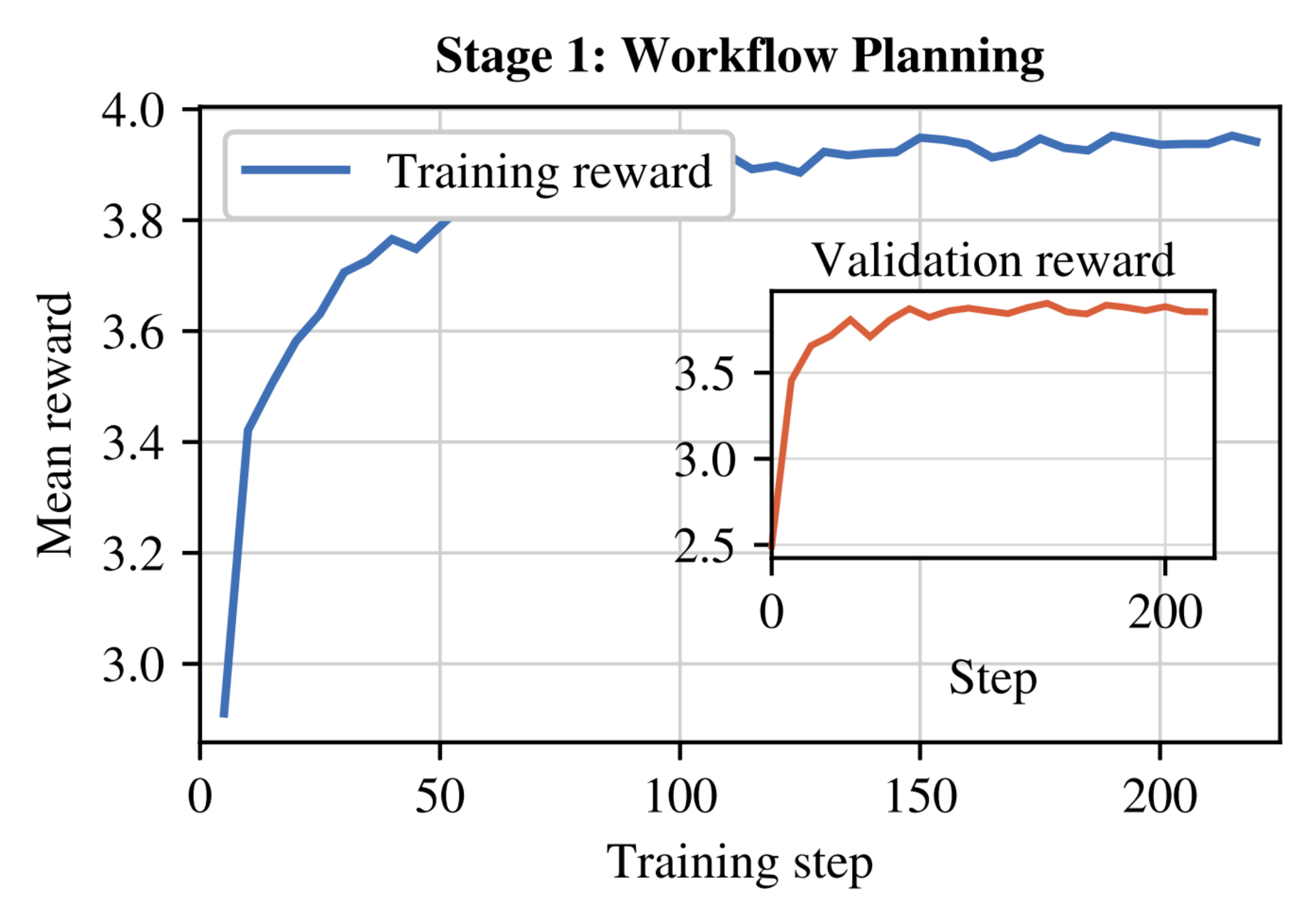}
  \caption{Stage~1 GRPO reward curve.}
  \label{fig:stage1_training_curve}
\end{figure}

\begin{figure}[tbp]
  \centering
  \includegraphics[width=\linewidth]{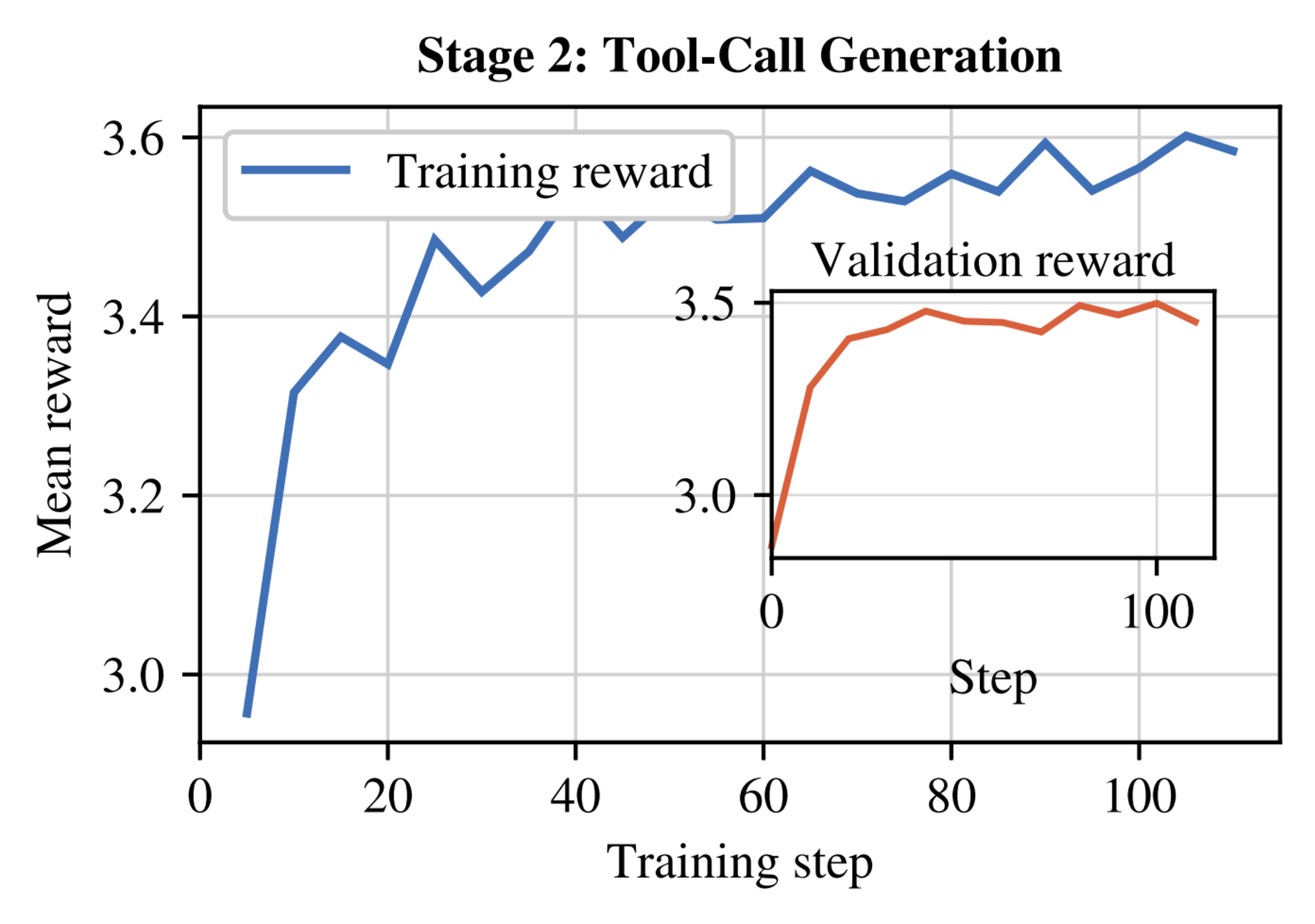}
  \caption{Stage~2 GRPO reward curve.}
  \label{fig:stage2_training_curve}
\end{figure}

Figures~\ref{fig:stage1_training_curve} and \ref{fig:stage2_training_curve} show the GRPO reward curves for the two training stages.

\FloatBarrier
\subsection{Case Study}
\label{app:case_study}
This subsection presents a successful case and a failure case to qualitatively examine the behavior of \methodname{}.

\subsubsection{Successful Case}

\paragraph{Task setup.}
We show a successful case using six tools from DailyLifeAPIs to examine whether the models can instantiate an explicitly requested workflow and preserve its cross-call dataflow.
The user query is:

\begin{quote}
\small
Archive the following project update as an audio recording.
Then transcribe the recording, convert the transcript into a PDF, and send the PDF to \texttt{alice@example.com}:

``The prototype review is complete, and deployment is scheduled for Friday.''
\end{quote}

The candidate tools and their corresponding FWG functions are summarized in Table~\ref{tab:successful_case_tools}.

\paragraph{Generated plans.}
Table~\ref{tab:successful_case} compares the tool-call plans generated by \methodname{} and ToolRL.
For compactness, ``update'' denotes the quoted project update, and ``email'' denotes the recipi ent address provided in the query.

\paragraph{Analysis.}
\methodname{} correctly instantiates the complete function-level workflow and preserves the dataflow
$p_1\rightarrow p_2\rightarrow p_3\rightarrow p_4$.
In contrast, ToolRL replaces PDF conversion with
\texttt{print\_document}, whose specified function is to print an existing document rather than convert it into a PDF.
Its final call also takes $\operatorname{out}(p_2)$ instead of $\operatorname{out}(p_3)$, bypassing the third call and sending the unconverted transcript.
This comparison shows how a locally plausible tool choice can violate the global workflow and produce an inconsistent argument dependency, whereas \methodname{} aligns concrete tool selection and source-traceable dataflow with the requested functional structure.

\subsubsection{Failure Case}

\paragraph{Task setup.}
We show a failure case using six tools from DailyLifeAPIs.
The user query specifies both a primary communication channel and a valid backup contact:

\begin{quote}
\small
Organize an online meeting named ``Prototype Deployment Review'' for Friday and take a note of the meeting details.
Email the note to Alice at \texttt{alice@example.com}.
Her backup phone number is \texttt{1234567890}.
\end{quote}

\paragraph{Generated plan.}
Table~\ref{tab:failure_case_plan} shows the plan generated by \methodname{}.
Direct arguments are copied from the query, while $\operatorname{out}(p_j)$ denotes a reference to the output of call $p_j$.
The candidate tools and their corresponding FWG functions are summarized in Table~\ref{tab:failure_case_tools}.

\paragraph{Analysis.}
\methodname{} correctly identifies the function-level workflow
Organization $\rightarrow$ Documentation $\rightarrow$ Transmission
and preserves the dataflow
$p_1\rightarrow p_2\rightarrow p_3$.
In particular, the \texttt{content} argument of the final call correctly references $\operatorname{out}(p_2)$.
However, the model instantiates the final Transmission function with \texttt{send\_sms}, even though the query requests delivery by email and provides the phone number only as a backup contact.
The correct tool should be \texttt{send\_email}, with \texttt{email\_address} set to \texttt{alice@example.com}.

This error occurs under an over-specified query that provides redundant but actionable contact information. Although the user explicitly requests email delivery, both the email address and the backup phone number provide sufficient arguments to instantiate a transmission tool.
The model appears to favor the phone number mentioned later in the query and consequently selects \texttt{send\_sms}.
This case suggests sensitivity to redundant information and its presentation order, rather than a failure to recover the global workflow or cross-call dataflow.

\begin{table*}[!t]
\centering
\small
\setlength{\tabcolsep}{6pt}
\renewcommand{\arraystretch}{1.15}
\begin{tabular}{@{}p{0.32\textwidth}p{0.22\textwidth}p{0.40\textwidth}@{}}
\toprule
\textbf{Tool Name} &
\textbf{FWG Function} &
\textbf{Description} \\
\midrule

record\_audio\_request
& Recording
& Records specified content as an audio recording. \\

record\_audio\_during\_call
& Recording
& Records audio during an ongoing call. \\

transcribe\_audio\_record
& Transcription
& Transcribes the content of an audio recording. \\

convert\_to\_pdf
& Conversion
& Converts an existing document into PDF format. \\

print\_document
& Output
& Prints a specified document rather than converting its format. \\

dispatch\_documents\_electronically
& Transmission
& Sends a document electronically to a specified destination. \\

\bottomrule
\end{tabular}
\caption{Candidate tools and their corresponding FWG functions in the successful case.}
\label{tab:successful_case_tools}
\end{table*}

\begin{table*}[t]
\centering
\small
\setlength{\tabcolsep}{5pt}
\begin{tabular}{@{}c p{0.18\textwidth} p{0.34\textwidth} p{0.34\textwidth}@{}}
\toprule
\textbf{Step} &
\textbf{Requested Function} &
\textbf{\methodname{}} &
\textbf{ToolRL} \\
\midrule
1 &
Recording &
record\_audio\_request
(content: update) &
record\_audio\_request
(content: update) \\

2 &
Transcription &
transcribe\_audio\_record
(content: $\operatorname{out}(p_1)$) &
transcribe\_audio\_record
(content: $\operatorname{out}(p_1)$) \\

3 &
Conversion&
convert\_to\_pdf
(document: $\operatorname{out}(p_2)$) &
print\_document
(document: $\operatorname{out}(p_2)$) \\

4 &
Transmission &
dispatch\_documents\_electronically
(document: $\operatorname{out}(p_3)$;
 destination: email) &
dispatch\_documents\_electronically
(document: $\operatorname{out}(p_2)$;
 destination: email) \\
\bottomrule
\end{tabular}
\caption{Comparison of the tool-call plans generated by \methodname{} and ToolRL in the successful case. Direct arguments are abbreviated by their semantic content, while $\operatorname{out}(p_j)$ denotes a reference to the output of call $p_j$.}
\label{tab:successful_case}
\end{table*}
\begin{table*}[!t]
\centering
\small
\setlength{\tabcolsep}{6pt}
\renewcommand{\arraystretch}{1.15}
\begin{tabular}{@{}p{0.31\textwidth}p{0.20\textwidth}p{0.43\textwidth}@{}}
\toprule
\textbf{Tool Name} &
\textbf{FWG Function} &
\textbf{Description} \\
\midrule

organize\_meeting\_online
& Organization
& Organizes an online meeting with a specified name and date. \\

take\_note
& Documentation
& Records supplied content as a note. \\

send\_email
& Transmission
& Sends specified content to an email address. \\

send\_sms
& Transmission
& Sends specified content to a phone number by SMS. \\

print\_document
& Printing
& Prints a specified document. \\

book\_hotel
& Reservation
& Books a specified hotel. \\

\bottomrule
\end{tabular}
\caption{Candidate tools and their corresponding FWG functions in the failure case.}
\label{tab:failure_case_tools}
\end{table*}
\begin{table*}[!t]
\centering
\small
\setlength{\tabcolsep}{5pt}
\renewcommand{\arraystretch}{1.15}
\begin{tabular}{@{}c p{0.17\textwidth}p{0.29\textwidth}p{0.42\textwidth}@{}}
\toprule
\textbf{Step} &
\textbf{FWG Function} &
\textbf{Selected Tool} &
\textbf{Arguments and Sources} \\
\midrule

1 &
Organization &
organize\_meeting\_online &
name: ``Prototype Deployment Review'' (\emph{direct});
date: ``Friday'' (\emph{direct}) \\

2 &
Documentation &
take\_note &
content: $\operatorname{out}(p_1)$ (\emph{reference}) \\

3 &
Transmission &
send\_sms &
phone\_number: ``1234567890'' (\emph{direct});
content: $\operatorname{out}(p_2)$ (\emph{reference}) \\

\bottomrule
\end{tabular}
\caption{Tool-call plan generated by \methodname{} in the failure case.}
\label{tab:failure_case_plan}
\end{table*}

\end{document}